\documentclass[11pt]{article}

\usepackage[preprint]{acl}

\usepackage{times}
\usepackage{twemojis}
\usepackage{array}      
\usepackage{tabularx}  
\usepackage{longtable} 
\usepackage{tabulary} 
\usepackage{natbib} 
\usepackage{multirow}   
\usepackage{makecell} 
\usepackage{booktabs}
\usepackage{latexsym}
\usepackage[most]{tcolorbox} 
\usepackage{amsmath}
\usepackage{enumitem}
\usepackage{amssymb}
\usepackage{subcaption}
\usepackage[dvipsnames]{xcolor}
\newcommand{\method}{{\Huge \twemoji{1f422}}\textcolor{ForestGreen}{TURTLE} }

\usepackage{xcolor}
\usepackage{listings}
\usepackage{caption}

\definecolor{codebg}{gray}{0.95}
\definecolor{kwpink}{RGB}{210,40,130}
\definecolor{strpurple}{RGB}{140,50,200}
\definecolor{linegray}{gray}{0.35}

\lstdefinestyle{pycode}{
  language=Python,
  backgroundcolor=\color{codebg},
  basicstyle=\ttfamily\small,
  keywordstyle=\color{kwpink},
  keywordstyle=[2]\color{kwpink},
  morekeywords=[2]{int,float,str,bool},
  stringstyle=\color{strpurple},
  numbers=left,
  numberstyle=\tiny\color{linegray},
  numbersep=8pt,
  frame=none,
  breaklines=true,
  breakindent=2em,
  columns=fullflexible,
  keepspaces=true,
  showstringspaces=false,
  xleftmargin=1.5em,
  aboveskip=0.5em,
  captionpos=b,  
  belowskip=0.5em,
}
\usepackage[T1]{fontenc}

\usepackage[utf8]{inputenc}

\usepackage{microtype}

\usepackage{graphicx}
\title{Evolving Self-tuning Benchmarks for Long-Horizon Tool Use}

\author{
 \textbf{Hyunji Alex Nam\textsuperscript{1\ddag}},
 \textbf{Michael Dennis\textsuperscript{2}},
 \textbf{Natasha Jaques\textsuperscript{23\dag}},
 \textbf{Minsuk Chang\textsuperscript{2\dag}},
\\
 \textsuperscript{1}Stanford University,
 \textsuperscript{2}Google Deepmind,
 \textsuperscript{3}University of Washington
\\
 \small{
   \textbf{Correspondence:} \href{mailto:hjnam@stanford.edu}{hjnam@stanford.edu}
 }
}

\begin{document}
\maketitle
\def\thefootnote{}\footnotetext{\textsuperscript{\dag}Equal advising. \textsuperscript{\ddag}Work done during internship at Google DeepMind.}
\addtocounter{footnote}{-1}
\begin{abstract}
For AI assistants to move beyond simple chat interfaces and reliably assist users with applications like Gmail, Venmo, and Slack, we need to improve their tool-use capabilities. However, existing tool use benchmarks are still limited and difficult to scale because they rely on human supervision or are too domain-specific. To make data generation scalable, we propose, \method(\textcolor{ForestGreen}{T}ool \textcolor{ForestGreen}{U}se \textcolor{ForestGreen}{R}elated \textcolor{ForestGreen}{T}asks with \textcolor{ForestGreen}{L}ong-Horizon \textcolor{ForestGreen}{E}volution), an Unsupervised Environment Design (UED) approach to evolve existing tasks using a proposer, a solver, and a feedback generator. The proposer is rewarded for \emph{difficult yet solvable} tasks based on the solver's empirical performance, and the feedback generator summarizes the solver's execution traces for the proposer to exploit. \method produces evolving, self-tuning test suites adaptable to different environment requirements, from supporting user simulators and dense document retrieval to interacting across multiple apps. Empirical results on AppWorld, $\tau$-Bench, and AutomationBench using Gemma 4 31B-it show that the evolved tasks target the desired difficulty range (e.g., a 165\% increase in the percentage of hard tasks on AppWorld), increase task diversity (in terms of reasoning and simultaneous tool interactions), and elicit a broader range of failure modes than the original tasks.
\end{abstract}

\section{Introduction} 


\begin{figure*}[t]
    \centering
    \includegraphics[width=\linewidth]{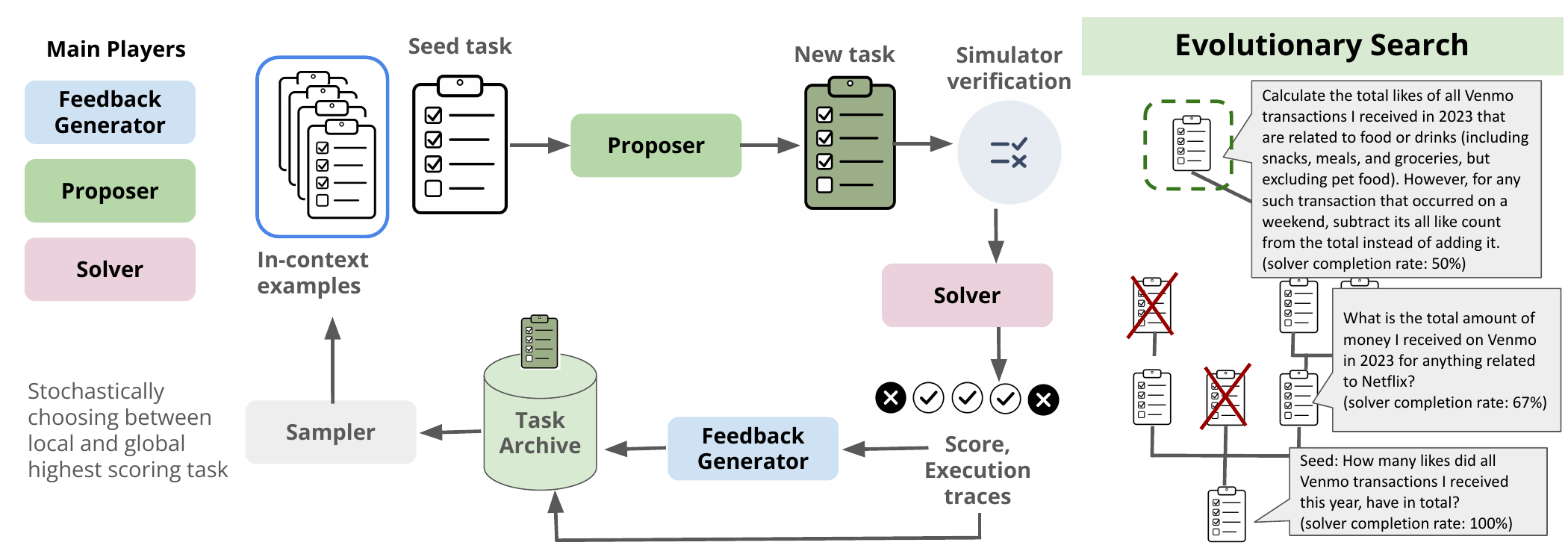}
    \caption{Even with the same underlying toolset, task difficulty can change depending on the instruction and the amount of reasoning and multi-step interaction required. Rather than building new benchmarks from scratch, \method evolves existing tasks using a task proposer, a solver, and a feedback generator LLM \emph{without human supervision.} During the evolutionary search, the proposer modifies the parent task, generating increasingly challenging but solvable tasks, while trivial or infeasible tasks are pruned based on their scores.}
    \label{fig:motivation_diagram}
    \vspace{-0.2cm}
\end{figure*}

As Large Language Models (LLMs) become more pervasive and integrated into our everyday lives and workflows, evaluating models across a wide range of scenarios is crucial for safe deployment. Especially for nascent applications like tool-use, evaluating models across diverse task suites is important for understanding their capabilities and vulnerabilities. However, existing tool use benchmarks are still limited and difficult to scale. Many existing benchmarks are domain-specific~\citep{merzouki2026factorybenchevaluatingindustrialmachine, shepard2026automationbench, yang2026programbenchlanguagemodelsrebuild}, or they rely on humans in the loop for task design and refinement~\citep{trivedi2024appworldcontrollableworldapps, shi2026tauknowledgeevaluatingconversationalagents, barres2025tau2benchevaluatingconversationalagents}. Furthermore, many new datasets are built from scratch and show little to no compatibility with the existing harnesses and testing setups of prior benchmarks, creating engineering overhead when incorporating new tests into an existing evaluation pipeline. Therefore, we need a scalable way to generate new tasks that test the robustness of models without adding to the engineering burden. Our key insight is that we \emph{do not need to reinvent the wheel}. Instead, we can reuse components of existing tasks to build new tasks. Existing tasks give us primitives and steps that we can recombine to produce a vast space of possible new tasks. By searching this space to target gaps in the solver's performance, we can still generate new tasks efficiently while ensuring task robustness.

We propose \method(\textcolor{ForestGreen}{T}ool \textcolor{ForestGreen}{U}se \textcolor{ForestGreen}{R}elated \textcolor{ForestGreen}{T}asks with \textcolor{ForestGreen}{L}ong-Horizon \textcolor{ForestGreen}{E}volution), an Unsupervised Environment Design (UED) approach~\citep{dennis2021emergentcomplexityzeroshottransfer, wei2026trainingsuperintelligentsoftwareagents, rutherford2024regretsinvestigatingimprovingregret} based on evolutionary search~\citep{ae_1993, mouret2015illuminatingsearchspacesmapping}. \method consists of a proposer, a solver, and a feedback generator, each responsible for new task generation, reward evaluation, and task reflection. Similar to UED and compositional environment design~\citep{gur2022environmentgenerationzeroshotcompositional}, we use a Proposer agent to automatically generate new tasks that target the solver's current level. We implement the proposer using an advanced evolutionary search operator, AlphaEvolve~\citep{novikov2025alphaevolvecodingagentscientific}. This allows our generation pipeline to incorporate highly capable models as the proposer---even closed-source ones---without requiring access to their weights, since evolutionary search handles gradient-free optimization. This is important given the large capability gap and varied behaviors observed across models of different sizes and release dates~\citep{marchand2026quantifyingfrontierllmcapabilities}. Evolutionary approaches have shown success in environment generation~\citep{faldor2025omniepicopenendednessmodelshuman}, persona generation~\citep{paglieri2026personageneratorsgeneratingdiverse, chopra2026cooperativesimulatorsgeneratingrealistic}, strategy discovery~\citep{wang2026discoveringdifferencesstrategicbehavior}, and red-teaming~\citep{samvelyan2024rainbowteamingopenendedgeneration}, but they remain under-explored for tool-use task generation.

The proposer receives an initial task and a toolset from open-source datasets and suggests mutations to the task based on the solver's prior execution traces and failure reflections. The modified task is then attempted by the solver $N$ independent times. To avoid tasks that are either too difficult or too easy, we design rewards based on the solver's completion rate, targeting a 50\% average completion rate, as in \emph{sampling for learnability}~\citep{rutherford2024regretsinvestigatingimprovingregret}, which maximizes the variance of the solver's success rate in 2D grid navigation environments. As a result, \method produces evolving, self-tuning benchmarks tailored to the current solver.


When evaluating the evolved tasks, we focus on three aspects: (1) the solver's empirical performance, (2) task diversity, and (3) elicited failure modes. Our empirical results span open-source benchmarks with different requirements, featuring user simulators~\citep{barres2025tau2benchevaluatingconversationalagents, shi2026tauknowledgeevaluatingconversationalagents}, documents for knowledge retrieval~\citep{shi2026tauknowledgeevaluatingconversationalagents}, 47+ simulated SaaS applications~\citep{shepard2026automationbench}, and extensive user databases and product catalogs modeled after real-world software applications like Amazon and Spotify~\citep{trivedi2024appworldcontrollableworldapps}. Our main results are reported with Gemma 4 31B-it~\citep{gemmateam2026gemma4technicalreport} as the solver, but we also include ablations with other models, such as Gemini 3.6 Flash~\citep{google2026gemini36flash} and Gemini 3.1 Flash Lite~\citep{google2026gemini31flashlite}, to show that \method adapts to different solver types. In summary, \method is a multi-agent evolutionary data-generation approach for eliciting long-horizon tool-use capabilities. It:\begin{itemize}
    \item Increases task difficulty when the original tasks are too easy (e.g., a 165\% increase in the proportion of hard tasks compared to the original AppWorld tasks for Gemma 4 31B-it)
    \item Increases the frequency of desirable task features, such as logical filtering and reasoning, by 29--212\% on AppWorld, and makes both semantic and functional changes to the task instructions;
    \item Can be used to diagnose solver vulnerabilities, broadening the coverage of (task feature $\times$ elicited error type) pairs from 39 to 44 out of 50 possible pairs.
\end{itemize}

\section{Related Work}
\paragraph{Tool-use benchmarks for LLMs.} Frontier models still struggle on long-context benchmarks requiring multi-step reasoning, such as FactoryBench~\citep{merzouki2026factorybenchevaluatingindustrialmachine}, AutomationBench~\citep{shepard2026automationbench}, ProgramBench~\citep{yang2026programbenchlanguagemodelsrebuild}, and $\tau$-bench (banking)~\citep{shi2026tauknowledgeevaluatingconversationalagents}. Despite their usefulness, these benchmarks are difficult to scale because they mainly provide static environments and tasks. Therefore, once models learn to solve problems in these datasets, the benchmarks cannot be reused to discover new tasks or model failures. 

Methods have been developed to generate new tasks. For example, Tool-R0~\citep{acikgoz2026toolr0selfevolvingllmagents} and GenEnv~\citep{guo2025genenvdifficultyalignedcoevolutionllm} operate only in text space to generate and verify tasks. However, LLM-simulated environments may suffer from hallucinations and inconsistencies~\citep{wang2024languagemodelsservetextbased}. Alternatively, \citet{lu2025toolsandboxstatefulconversationalinteractive, barres2025tau2benchevaluatingconversationalagents, yao2024taubenchbenchmarktoolagentuserinteraction, shi2026tauknowledgeevaluatingconversationalagents, trivedi2024appworldcontrollableworldapps, esfandiarpoor2025themcpcompanycreatinggeneralpurposeagents} build tool-interactive sandboxes and database states that more closely resemble real-world applications. However, these are challenging to scale, as they require humans to refine tasks or evaluation environments for robustness. Furthermore, they are often tailored to specific domains (e.g., $\tau$-bench for travel and retail~\citep{yao2024taubenchbenchmarktoolagentuserinteraction}, banking~\citep{shi2026tauknowledgeevaluatingconversationalagents}, and telecom~\citep{barres2025tau2benchevaluatingconversationalagents} are all separate benchmarks). Other methods rely more heavily on prompting for environment design and task discovery, such as EnvScaler~\citep{song2026envscalerscalingtoolinteractiveenvironments} and Agent World Model~\citep{wang2026agentworldmodelinfinity}. Agent-World~\citep{dong2026agentworldscalingrealworldenvironment} uses prompts to adjust task difficulty with hand-designed task features, such as the number of reasoning steps and the number of unique tools used per task. Validating task feasibility remains challenging in automated task generation. To guarantee solvability, ToolGrad~\citep{zhou2026toolgradefficienttoolusedataset} reverses the task-generation process by first constructing the ground-truth tool-call sequence, then writing a matching natural language instruction for the given sequence. \method is different from prior approaches in two ways: (1) evolutionary: using gradient-free evolutionary optimization to incorporate advanced models, and (2) self-tuning: minimizing human oversight in task design and adapting to particular solvers.

Our work shares the motivation of \citet{cai2025autoforge, fang-etal-2026-towards}, which use existing tool-use datasets to automate task generation. However, they rely on state-dependency graphs and random walks, which can limit the flexibility of task generation, and building graphs become challenging as the number of tools increases and their inter-dependencies grow complex. \method combines the best of both approaches by leveraging existing benchmarks and verified API implementations, while evolving them to adjust their difficulty based on the solver's performance.

\paragraph{LLM-powered evolutionary search.} Evolutionary algorithms are gradient-free optimization inspired by biological evolution~\citep{ae_1993, mouret2015illuminatingsearchspacesmapping}. They iteratively mutate a population of programs and select candidates based on their fitness scores, while allowing fitter individuals to reproduce more often than less fit ones. Incorporating large language models into evolutionary search has proven successful in scientific and mathematical optimization~\citep{novikov2025alphaevolvecodingagentscientific, codeevolve}, red-teaming~\citep{yuan2026agenticredevolvingagenticsystems}, and social applications using conversational assistants, such as generating diverse user personas~\citep{chopra2026cooperativesimulatorsgeneratingrealistic, paglieri2026personageneratorsgeneratingdiverse} and strategy discovery~\citep{wang2026discoveringdifferencesstrategicbehavior}.  AlphaEvolve~\citep{novikov2025alphaevolvecodingagentscientific} and OpenEvolve~\citep{openevolve} are among the popular frameworks for orchestrating evolutionary search using LLMs.


\section{Method} \method consists of three key LLM agents: (1) a \textbf{proposer} implemented with AlphaEvolve; (2) a \textbf{solver}, which attempts the new task yielding fitness scores; and (3) a \textbf{feedback generator}, which identifies the main reason for failure based on the solver's execution traces and the ground-truth solution. We leverage multi-agent dynamics to propose, evaluate, and refine task mutations according to the solver's problem-solving competence. 

After the evolutionary search is run for a fixed number of steps, we add programs with the highest completion rate variance (including ties) per parent task from the original dataset to the final dataset $\mathcal D_\text{\method}$. We then apply a filtering step with an LLM judge (Gemini 3.6 Flash) to remove tasks with trivial solver failures, such as formatting mismatches or paraphrases of the same content, which can be attributed to ambiguous or underspecified instructions (see judge prompt in Appendix Fig.~\ref{appendix:trivial_solver_failure}).

\subsection{Evolutionary Task Proposer}
\textbf{Task structure.} A task is presented as a Python program, and the proposer is instructed to act as a programmer. In particular, on AppWorld, we provide minimal instructions to explore how AlphaEvolve discovers task features on its own (proposer instruction prompts are included in Appendix~\ref{appendix:task_proposer_prompts}).

Depending on the features of the initial task, the proposer is given different items specifying the task. For example, in AppWorld, a task is defined by: \textbf{(1) the initial database state}, which specifies the user's account information, friends list, message history, Amazon product catalogs etc.; \textbf{(2) the original task instruction in natural language} (e.g., ``\emph{Give me a comma-separated list of the six most-played EDM songs from my Spotify playlist.}"); \textbf{(3) the ground-truth solution}, which shows the sequence of API calls and their arguments; and \textbf{(4) optionally, evaluation criteria}. If a task changes only part of the database state, the evaluation criteria are set automatically to the oracle final state obtained by executing the ground-truth solution in a simulated environment, so the proposer does not need to specify them. If enumerating all state changes is too expensive, the proposer additionally specifies which parts of the database must be inspected, and the solver is evaluated only on the changes made to those parts.


\noindent \textbf{Task mutation.} During each evolutionary step, the proposer is given a parent task and five additional tasks sampled from the archive, stochastically chosen between the best global and local tasks (highest-scoring per metric). Each task is represented by its instruction text, a corresponding score, and a summary by the feedback generator. The proposer uses this information to propose modifications to the parent task that can maximize the fitness function. Unlike prior work that uses RL post-training to update the proposer~\citep{acikgoz2026toolr0selfevolvingllmagents, guo2025genenvdifficultyalignedcoevolutionllm}, we frame task generation as a search problem seeded with a parent task and use AlphaEvolve to discover variants of the original task. 

\begin{lstlisting}[
  caption={Appworld task as a program.},
  label={lst:template}
]
def task():
    return { 
        db_changes: # initial database state,
        full_description: # user account information (e.g., acces tokens, emails, tool-discovery instruction),
        instruction: # natural language instruction,
        ground_truth_calls: # ground truth API sequence,
    }
\end{lstlisting}
\vspace{-0.2cm}

\noindent \textbf{Task verification.} Before sending a new task to the solver, we execute the proposed ground-truth solution in the simulated environment to verify there are no system errors. This step filters out tasks with inconsistent state dependencies (e.g., asking to send a message to a friend who is not in the contact list). If the proposer specifies evaluation criteria, we verify that the ground-truth solution achieves a perfect score under those criteria.

\subsection{Solver (Task Difficulty Evaluator)}\label{sec:solver} A new task is evaluated based on the solver's empirical performance. In particular, we reward tasks that are solvable but not consistently solved perfectly. Each task is attempted $N$ times, and an attempt scores 1 if the solver fully completes the task; otherwise 0 (partial credits are allowed in AutomationBench~\citep{shepard2026automationbench} to avoid reward sparsity on harder domains). To avoid penalizing trivial string mismatches, we additionally prompt the LLM judge to evaluate semantic state matching using the original task instruction and the observed versus expected final states.

\noindent \textbf{How to avoid reward hacking.} Some works on multi-agent task generation leverage zero-sum adversarial dynamics, where the proposer attempts to minimize the solver's performance, in order to ensure robustness~\citep{liu2026chasingmovingtargetsonline}. However, in the case of tool-use task generation this objective is badly posed. Adversarial dynamic between the proposer and the solver can lead to reward hacking, since the proposer may be incentivized to generating ambiguous or partially solvable instructions. Therefore, to ensure the tasks are feasible, we check that they are solved perfectly in at least one of the solver attempts. 


\noindent \textbf{Fitness function.} Our main objective is to generate long-horizon tasks that can reveal different model failures. To this end, we construct a multi-objective reward function based on the solver's completion rate. We define $c$ as the solver's average completion rate across $N$ independent attempts, and $L$ as the average solver's trajectory length. We propose the following six metrics:\begin{itemize}
\item Target difficulty reward: $\exp \{-4 \ln(10) (c - 0.5)^2\}$ how close $c$ is to the target value of 0.5; a bell-curve with a maximum of 1 at $c=0.5$ and a minimum of 0.1 at the endpoints;
\item Sampling for learnability~\citep{rutherford2024regretsinvestigatingimprovingregret}: $c (1- c)$;
\item Trajectory length bonus $\times$ relaxed difficulty threshold: $\max\{1, L\} *  I\{0.3 \leq c \leq 0.7\}$;
\item Trajectory length bonus $\times$ strict difficulty threshold: $\max\{1, L\} *  I\{0.4 \leq c \leq 0.6\}$;
\item Trajectory length bonus $\times$ sampling for learnability: $\max\{1, L\} * c (1 - c)$;
\item Trajectory length bonus $\times$ target difficulty: $\max\{1, L\} *  \exp\{-4 \ln(10) (c - 0.5)^2\}$.
\end{itemize}
To encourage more complex tasks, we reward the proposer for generating longer tasks that the solver is still able to complete. To prevent the proposer from trivially extending the ground-truth task length with unnecessary tool calls that do nothing (a reward hacking behavior we initially observed), we base these rewards on the length of the solver's successful trajectory.

During the evolutionary search, the proposer stochastically chooses between global optimality (the highest-scoring task on average across six metrics) and local optimality (the highest-scoring task for one of the six metrics) to balance exploration and exploitation. When suggesting modifications, the proposer reviews the parent task, along with five sampled in-context examples, and their corresponding textual and numeric performance summaries, and is instructed to optimize the average fitness score based on the individual score report.



\subsection{Feedback Generator} One key advantage of LLM-powered evolutionary search is that the task proposer can use natural-language feedback, in addition to fitness scores, to guide its search for better performing mutations. To leverage this strength, we incorporate a feedback generator LLM into the evolutionary search. For tasks with unsuccessful solver attempts, the feedback generator outputs a short explanation of what went wrong by comparing the solver's outcome against the ground-truth solution and the provided task instruction. This feedback is stored in the task archive alongside the solver's execution traces and provided to the proposer during mutation.

\section{Experiment} 

\method is applied to four different benchmarks: AppWorld~\citep{trivedi2024appworldcontrollableworldapps}, $\tau$-telecom~\citep{barres2025tau2benchevaluatingconversationalagents}, $\tau$-banking~\citep{shi2026tauknowledgeevaluatingconversationalagents}, and AutomationBench~\citep{shepard2026automationbench}. More details of each environment in Appendix~\ref{appendix:dataset}.

\noindent \textbf{AppWorld} is built on 4,571 executable APIs across widely used, everyday software applications. Each task is associated with a particular user and their personal data (e.g., login credentials, transaction history, and product search history), and the agent completes tasks as the user's personal assistant.

\noindent \textbf{$\tau$-bench (banking \& telecom)} requires the model to operate as a customer service agent in a dual-control, multi-turn dialogue setting with a (LLM) simulated user. Certain tool calls can be executed only by the user, so the agent must guide the user through the process step by step. $\tau$-banking is designed to test the agent's document retrieval and long-context reasoning (consisting of 698 documents spanning 21 product categories and financial topics). We test the solver with two of the retrieval conditions presented in~\citet{shi2026tauknowledgeevaluatingconversationalagents}: BM25 and golden retrieval.

\noindent \textbf{AutomationBench} uses 47 simulated SaaS applications to model realistic workflows involving tools such as Gmail, Google Sheets, Slack, Salesforce, Zendesk, Jira, and HubSpot. The agent needs to navigate through business objectives and guardrails to solve tasks spanning domains such as Finance, HR, Marketing, Operations, Sales, and Customer Support. 

Evolutionary search is implemented with AlphaEvolve~\cite{novikov2025alphaevolvecodingagentscientific}. Tasks are evolved for 1,000 steps for AppWorld and 500 steps for the other datasets, and every new task is attempted 10 times. \method{} is equally applicable to both closed- and open-weight models.

Main empirical results are reported using Gemma 4 31B-it~\citep{gemmateam2026gemma4technicalreport} as both the solver and the feedback generator. We also conduct ablations with two proprietary models, Gemini 3.6 Flash~\citep{google2026gemini36flash} and Gemini 3.1 Flash Lite~\citep{google2026gemini31flashlite}.

\section{Results}

\subsection{\method adjusts difficulty to solver} Data scaling can be done trivially if tasks are padded with unnecessary or repeated tool calls. To show the usefulness of evolving, self-tuning benchmarks, we compare the difficulty of the original ($n=106$) and new tasks ($n=735$) evolved from AppWorld. We assess difficulty in two ways: (1) the average completion rate; and (2) the LLM-judge evaluation using a task difficulty rubric from ~\citet{trivedi2024appworldcontrollableworldapps}. In their work, humans rate tasks on a scale of 1 (easy) to 3 (hard) based on the amount of interaction and reasoning required. We adapt their prompt to enable a LLM-judge using Gemini 3.6 Flash (prompt in Appendix Fig.~\ref{appendix:llm_difficulty_prompt}). 

\begin{figure}[t]
\centering
\includegraphics[width=\linewidth]{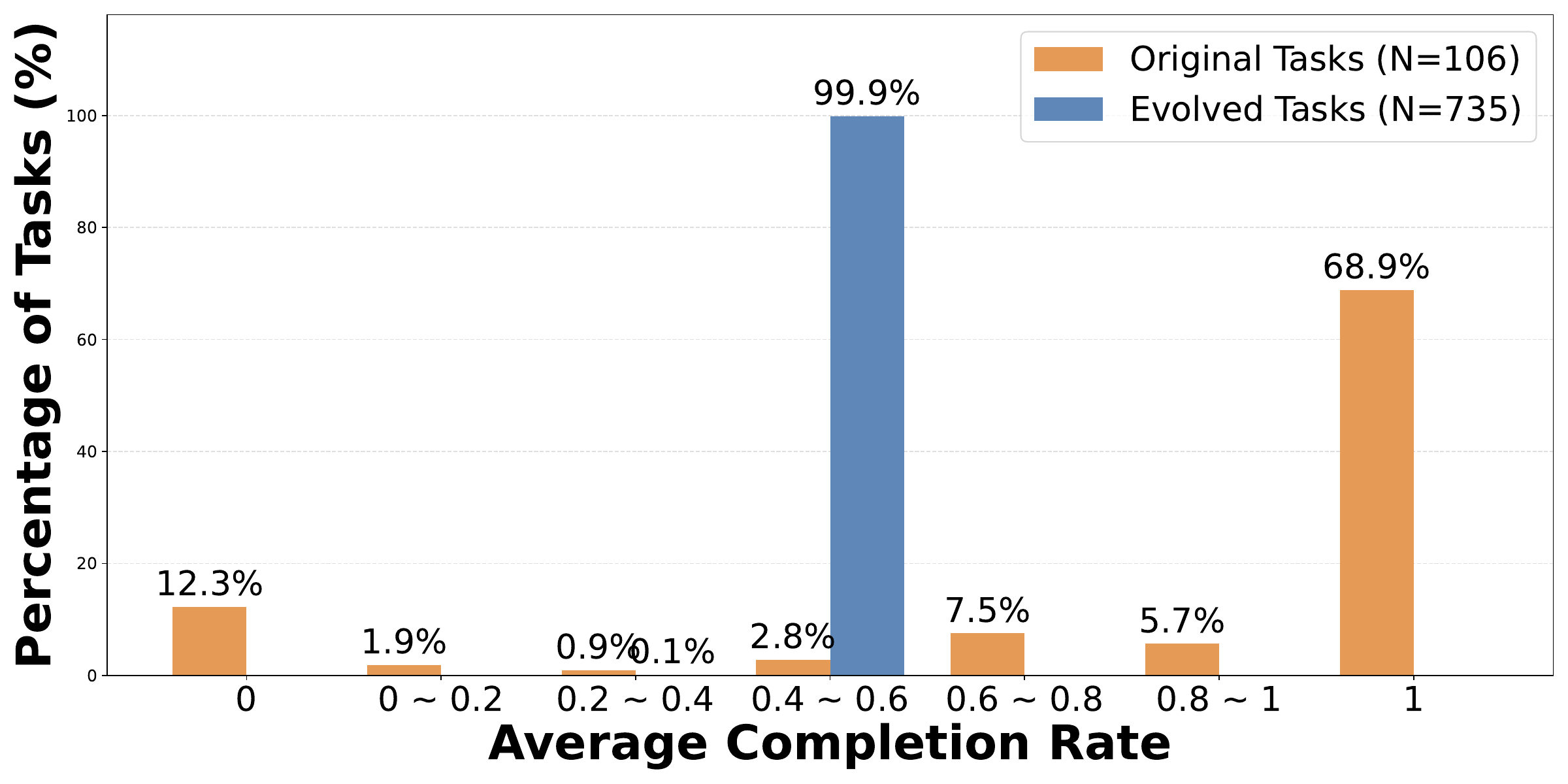}
\caption{Distribution of average completion rates on the original AppWorld (orange) vs \method (blue) tasks.}
\label{fig:difficulty-distribution}
\vspace{-0.2cm}
\end{figure}


\begin{figure}[t]
\centering
\includegraphics[width=\linewidth]{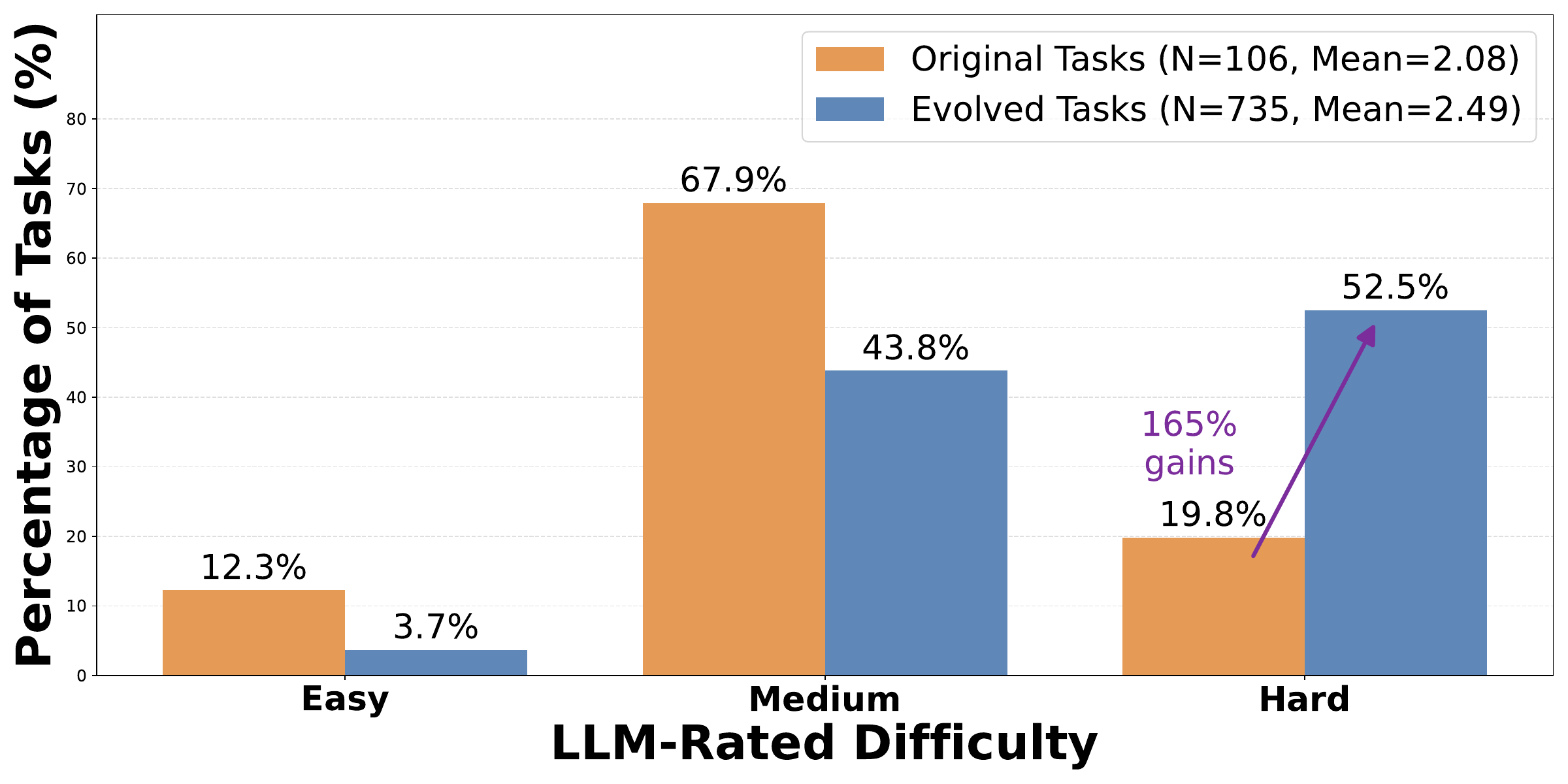}
\caption{Distribution of LLM-evaluated difficulty levels on the original AppWorld vs \method tasks.}
\label{fig:llm_difficulty_distribution}
\vspace{-0.5cm}
\end{figure}

\noindent \textbf{Solver's empirical performance.} Fig.~\ref{fig:difficulty-distribution} shows that the initial AppWorld task difficulty distribution, as measured by completion rate, is largely bimodal, with 69\% of tasks achieving perfect completion and 12\% always failing. After ~\method, nearly all tasks lie in the 40--60\% average completion rate range. 
Ablations with different solver models and initial task datasets are reported in Appendix Figs.~\ref{appendix:appworld_completion_rates} and \ref{appendix:fig_task_completion_rates}~\footnote{In $\tau$-telecom, we obtain 299 tasks from 84 parent tasks; in $\tau$-banking (BM25), we obtain 82 from 17; in $\tau$-banking (golden retrieval), we obtain 178 from 24; and on AutomationBench, we obtain 23 from 16. Gemini 3.1 Flash Lite yields 141 tasks from 64 seeds in AppWorld, while Gemini 3.6 Flash yields 22 from 9.}. Across different setups, \method{} consistently generates tasks that largely fall into the 40-60\% average completion rate, regardless of the original task difficulty. This shows that the self-evolving benchmarks target the solver's learning frontier, i.e., identifying solvable problems that the solver sometimes trips over.

\noindent \textbf{LLM-judge evaluation.} The original AppWorld tasks are often too easy for the Gemma 4 31B-it solver, and the LLM-judge evaluations in Fig.~\ref{fig:llm_difficulty_distribution} further confirm that 80\% of the original tasks are considered either easy or medium. In contrast, \method-generated tasks are evaluated as 44\% medium and 53\% hard, which is a 165\% increase in the proportion of hard tasks compared to the original dataset (i.e., 20\%). 

\subsection{Emergent features from self-tuning benchmarks}
When evolving tasks in AppWorld, we give minimal instruction to the task proposer on how to mutate tasks. The only signals the proposer receives are the fitness scores based on the solver's performance and summaries by the feedback generator. Despite minimal guidance, \method{} discovers new task features that can make the tasks harder. 

\begin{figure}[t]
\begin{tcolorbox}[
    colback=white,
    colframe=blue!70!black,
    coltitle=white,
        title=Hand-picked examples of evolved task features,
    fonttitle=\bfseries
]
\small
\textbf{\# Applying constraints \& filtering logic} \\ 
Original task: \emph{Download all the songs from my Spotify album library that I have liked.} \\
New task: \emph{Download all the songs from my Spotify album library that I have liked and were released in 2022 or later.} \\ \\ 
\textbf{\# Introducing multi-step tool calls} \\
Original task: \emph{Download all the songs from my Spotify album library that I have liked.} \\ 
New task: \emph{Download all the songs from my Spotify album library that I have liked. For each of these songs, create a
note in my Simple Note account with the song's title as the title and the string 'Genre: , Duration: s' as the
content. Tell me the total number of songs downloaded and the total number of notes created.} \\ \\ 
\textbf{\# Hiding variables behind other tool calls} \\ 
Original task: \emph{The last Venmo payment request I sent to Cory was an accident and they approved it. Send them the
money back.} \\ 
New task: \emph{Find my roommate's email and refund them for the 'movies' payment they made. Use 'Accidental Request
Refund' as the description.}
\end{tcolorbox}
\caption{Features discovered during \method.}
\vspace{-0.5cm}
\label{fig:task_feature_examples}
\end{figure}

\begin{figure}[h]
    \centering
    \includegraphics[width=\linewidth]{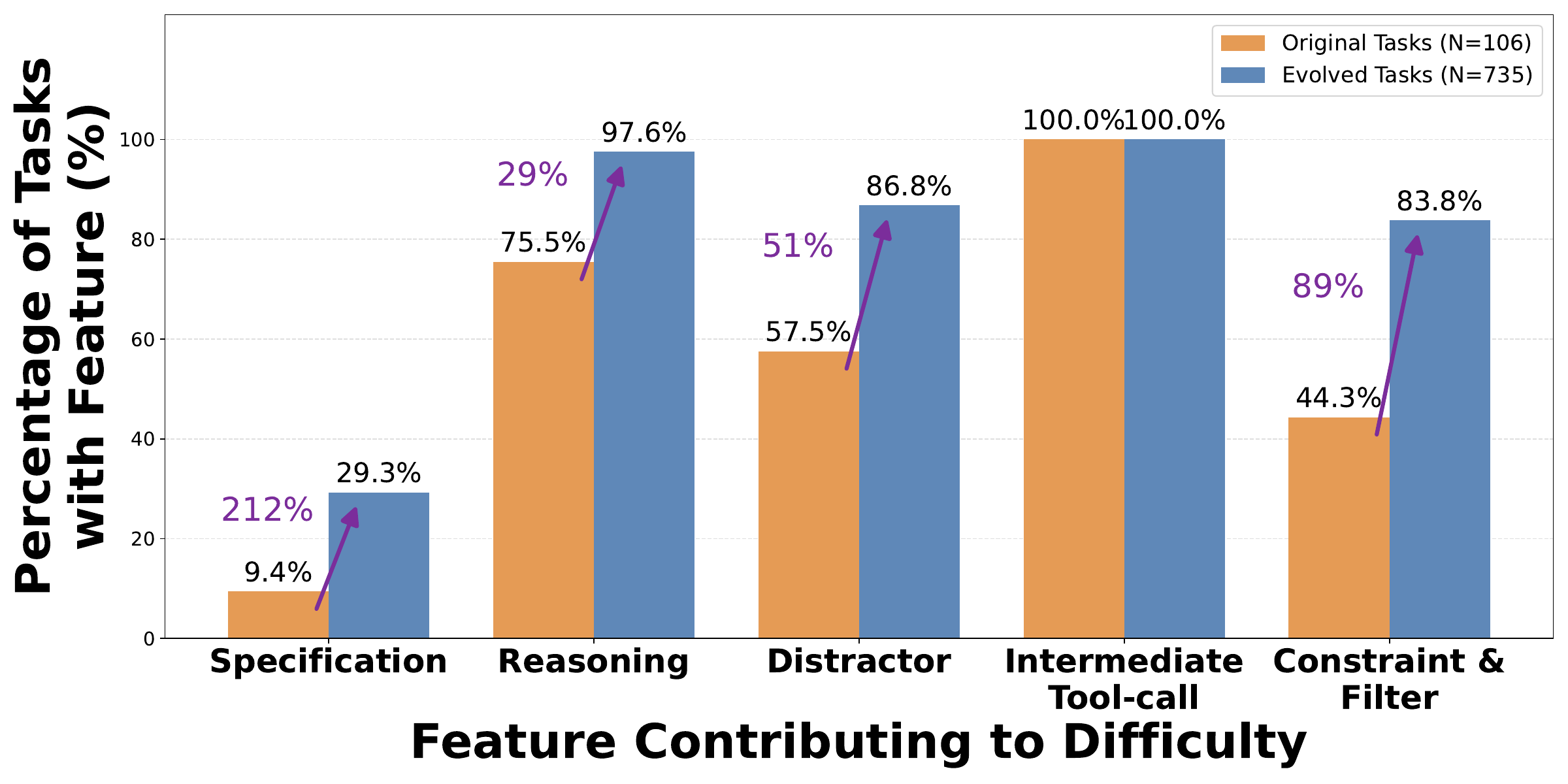}
    \caption{Frequency of task features in the original AppWorld (orange) vs \method (blue) datasets.}
    \label{fig:task_feature_distribution}
    \vspace{-0.5cm}
\end{figure}


We build a taxonomy of task features: (1) specification, (2) reasoning, (3) distractors, (4) intermediate tool calls, and (5) constraints and filters (see Appendix Table~\ref{appendix:task_feature_taxonomy} for definitions and examples) and use an LLM judge (Gemini 3.6 Flash) to annotate each task with the observed feature for comparison in Fig.~\ref{fig:task_feature_distribution}. Intermediate tool calls are always present in both the original and evolved task sets, but the percentage of all other task features increase as large as 212\% after \method. 

\begin{figure}[htbp]
    \centering
    \includegraphics[width=\linewidth]{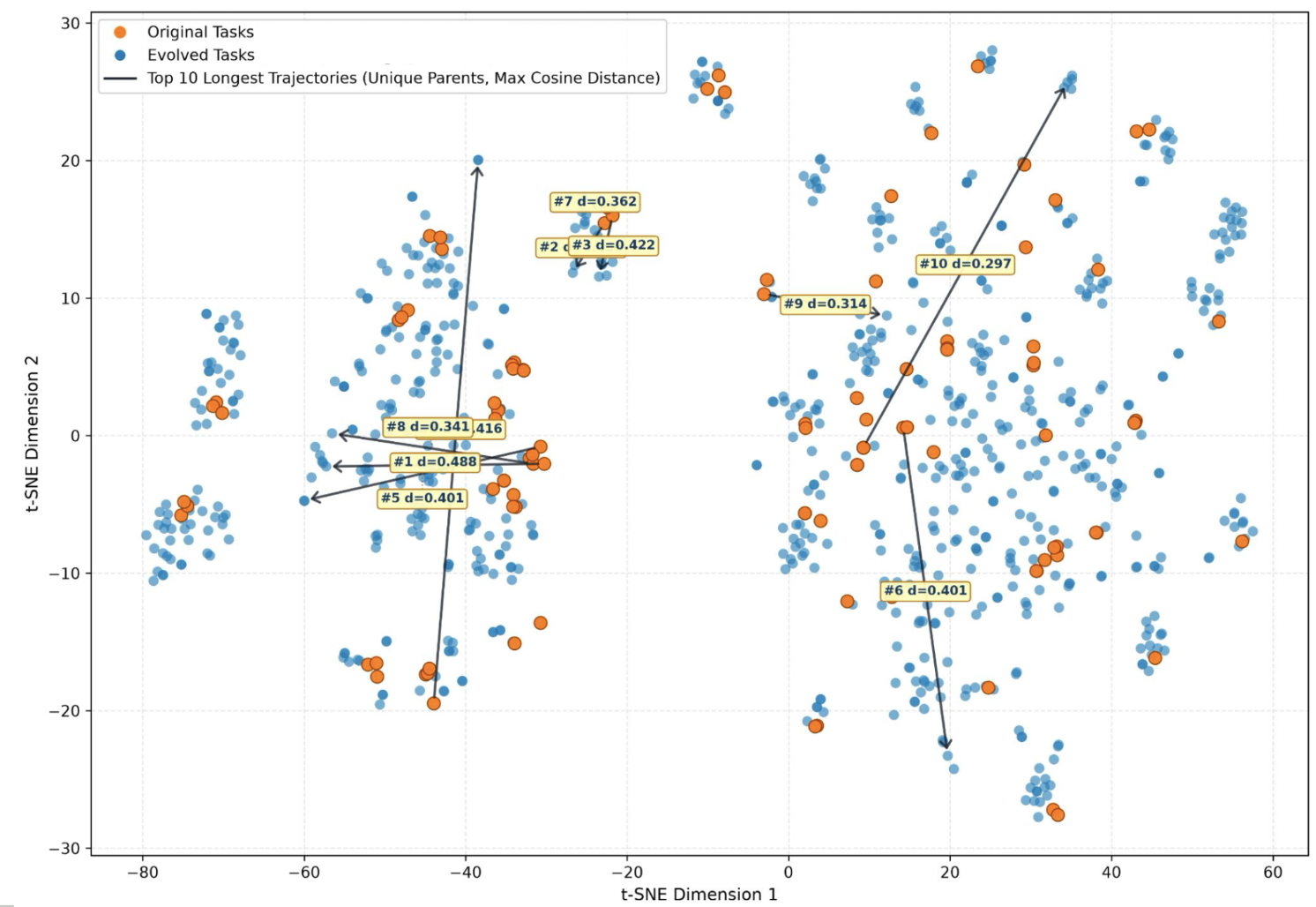}
    \caption{t-SNE visualization of task instruction embeddings in orange (original) and blue (\method). The arrows show the top 10 original-evolved task pairs with the largest cosine distances. \method generates new tasks that are semantically different from the original tasks, expanding the search space.}
    \label{fig:task_semantic_embedding}
\vspace{-0.5cm}
\end{figure}

\subsection{\method enhances diversity} Another desideratum for task evolution is whether the tasks are modified in semantically and functionally meaningful ways beyond making natural-language or numeric perturbations to the original task instructions.

\noindent \textbf{Text embeddings of task instructions.} To examine whether \method{} makes semantically meaningful changes to the original dataset, we embed the original and generated task instructions using Gemini Embedding 2~\citep{shanbhogue2026geminiembedding2native} (with 768 embedding size) and present a t-SNE visualization~\citep{JMLR:v9:vandermaaten08a} of the embeddings in Fig.~\ref{fig:task_semantic_embedding} (This figure shows results from experiments with the Gemma solver on AppWorld; more examples in Appendix Fig.~\ref{appendix:tsne_viz}). The top 10 largest cosine distances range from 0.30 to 0.49, with the largest change between the original instruction: \texttt{Send \$68 privately on Venmo to the person with this phone number 2114992604}, and the evolved instruction, involving multi-step tool calls and reasoning about state dependencies: \texttt{Find my birth month from my profile. Then, find all my coworkers who were also born in that month and live in Seattle. Send \$75 to each of them privately on Venmo. For the one who is oldest, use the note `To the most senior birthday buddy!'. For all the others, use `Happy birthday [First Name] buddy!', replacing [First Name] with their actual first name (e.g., `Happy birthday Sherry buddy!'). Finally, send an email to the oldest coworker with the subject `Happy Birthday!' and body `Hope you have a great one!'.} In this example, the evolved instruction requires the solver to first find each user’s birth month, send messages to all users who satisfy the condition, and customize each message based on the user’s attributes, which are missing from the original instruction. We also provide examples of a task lineage in Appendix Fig.~\ref{fig:task_lineage1}, ~\ref{fig:task_lineage2}, ~\ref{fig:task_lineage}, showing how the proposer gradually adds new steps and variables to increase the task difficulty.

\begin{figure*}[htbp]
    \centering
    \includegraphics[width=\linewidth]{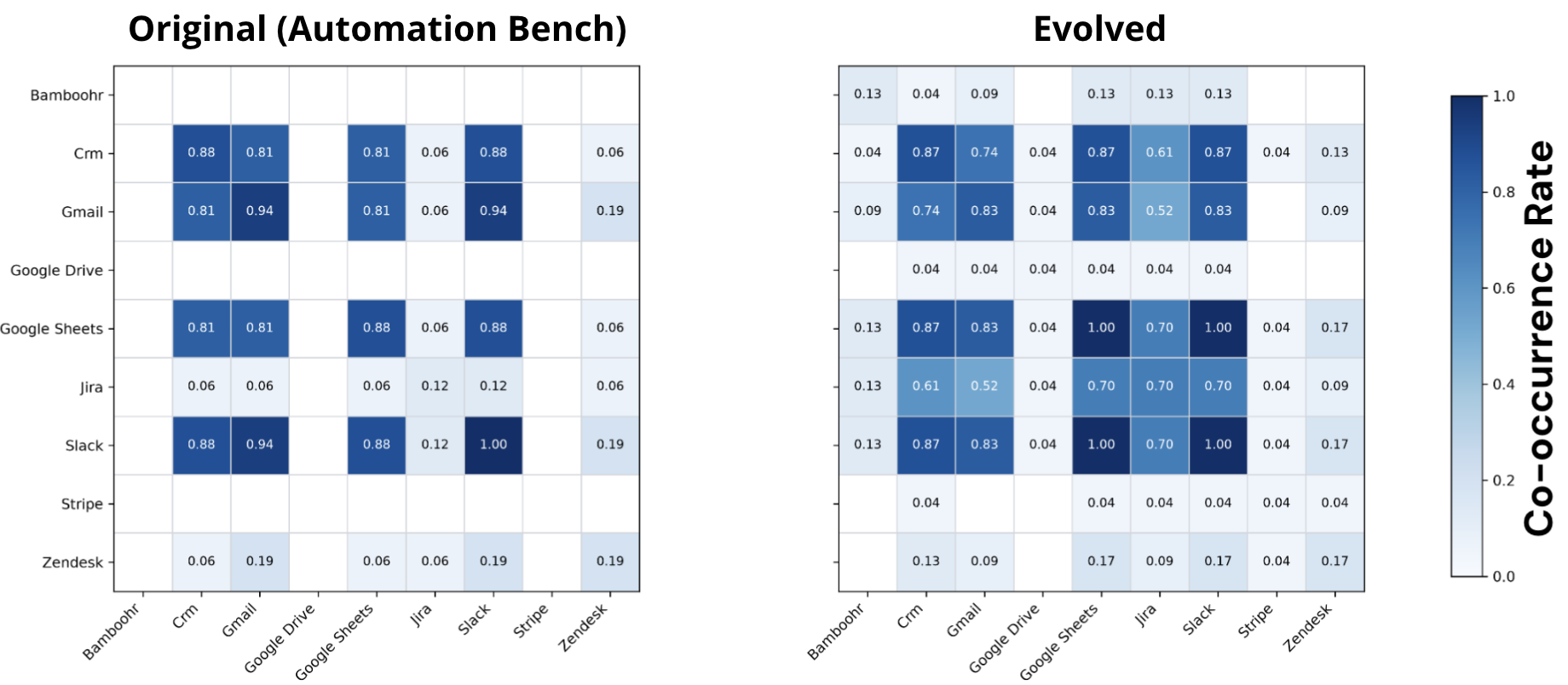}
    \caption{Co-occurrence of apps between original (left) and \method (right) tasks in AutomationBench. Diagonal entries indicate the percentage of tasks that use that particular app rather than co-occurrence of two apps.} \label{fig:app_coocurrence}
    \vspace{-0.2cm}
\end{figure*}

\noindent \textbf{Diversity of simultaneous tool calls.} We also observe changes in task requirements (i.e., the number and types of software applications used) after applying \method{} to AutomationBench. Initially, all 16 original tasks require four apps, but after evolution, 52\% of the 23 evolved tasks require five apps and 13\% require six. The changes are not limited to the number of apps used. Fig.~\ref{fig:app_coocurrence} shows the frequency of app co-occurrences in the original and evolved tasks. The number of unique app pairs required by each task also increases from 15 to 30 (out of 36 possible unordered pairs), and the evolved tasks now require using new apps like Bamboohr, Google Drive, and Stripe. \method{} makes substantive changes to the underlying tool interactions, rather than modifying instructional styles or function parameters in a templated way.

\subsection{\method for failure diagnosis}\label{sec:failure_analysis} \method can elicit new model failure modes, whereas the original tasks are often too easy to highlight solver vulnerabilities. For example, in the AppWorld datasets, we observe that Gemini 3.1 Flash Lite and Gemini 3.6 Flash sometimes struggle with ambiguous instructions or responses and fail at non-tool-calling related reasoning steps, while Gemma 4 31B-it often fails to infer the user's intent when following complex instructions.
\begin{itemize}[noitemsep, topsep=0pt]
\item \textbf{Original task: } \emph{How many people follow the artist of the currently playing song on Spotify?} (solved perfectly)
\item \textbf{New task: } \emph{Check the currently playing song on Spotify. I need a passcode. The passcode is the song's duration minus the song\_id, multiplied by the number of artists, and then concatenated with the last word of the song's title in uppercase. For example, if duration is 300, song\_id is 100, there are 2 artists, and the title is 'Hello World', the passcode is '400WORLD'. What is the passcode?}
\item \textbf{Gemini 3.1 Flash Lite }  interprets the artist's name, ``Seraphina Dawn", as two artists and multiplies incorrectly. This shows the model lacks the commonsense understanding needed to resolve ambiguity in the text response. 
\item \textbf{Original task: } \emph{Send the following phone message to my siblings and roommates, who do not have a venmo account, ``Get on venmo please!".} (solved perfectly)
\item \textbf{New task: } \emph{I need to contact people I live with and my brothers, but not my coworkers. Find out if they use Venmo. For those who don't, text them `Get on venmo please!'. For people I live with that are on Venmo, send them a reminder to `Pay me back!'. Don't message my brothers if they're on Venmo.}
\item \textbf{Gemma 4 31B-it } sends a Venmo payment request with a hallucinated amount of \$1 instead of sending a text reminder, as requested.
\end{itemize}

\begin{figure*}[htbp]
    \centering
    \includegraphics[width=\linewidth]{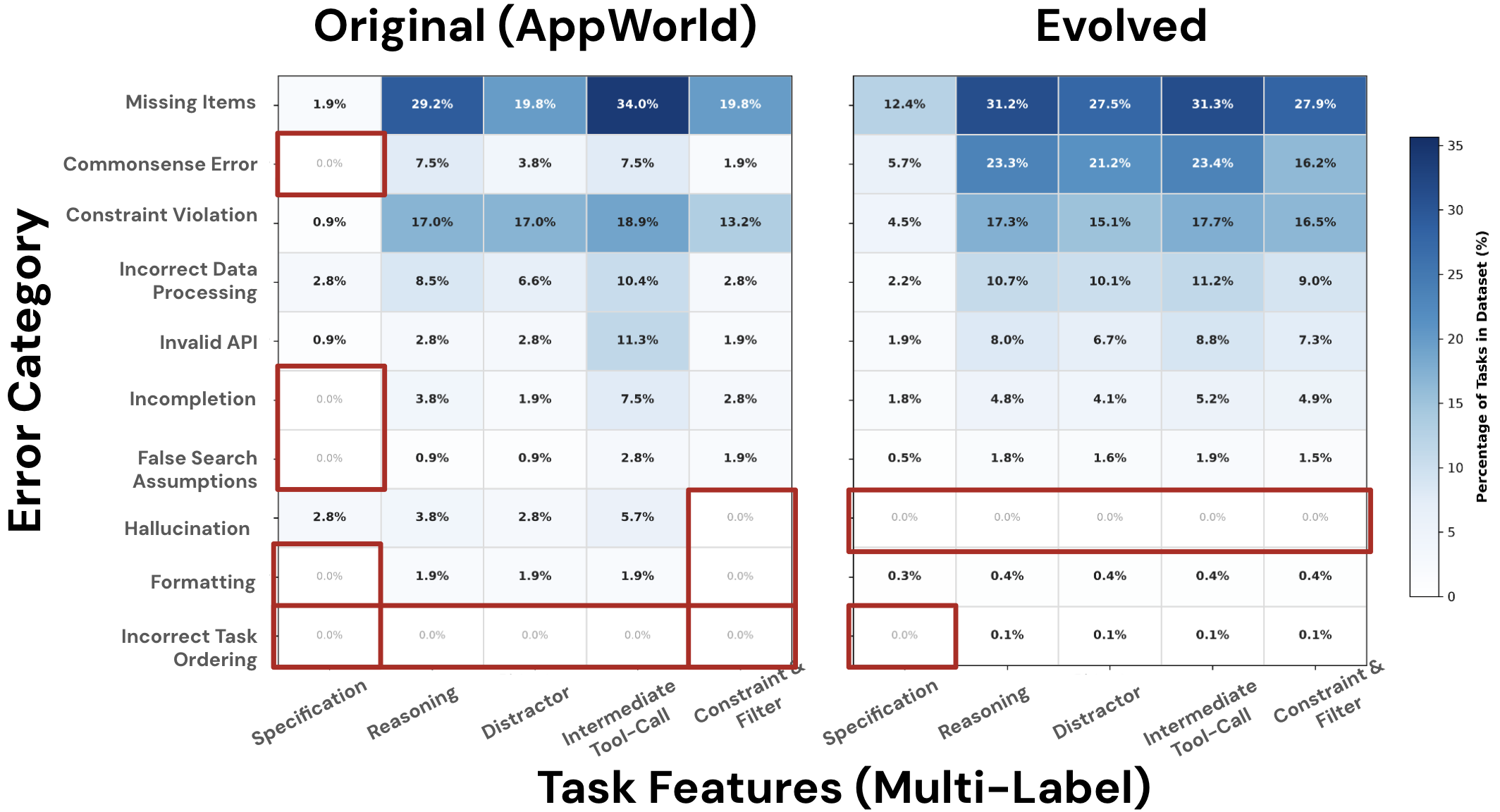}
    \caption{Task features versus error categories for the original (left) and evolved (right) tasks. Darker cells indicate higher frequency and cells with 0 errors are highlighted in red. Tasks with 100\% completion rates are excluded from the numerator since they do not elicit any failure, but they remain in the denominator.} \label{fig:failure_taxonomy}
\end{figure*}

We also build a taxonomy of solver failure modes (Appendix Table~\ref{appendix:failure_taxonomy}) and use Gemini 3.6 Flash as the judge to identify the primary failure reason for each task with any observed failure. For both the original and \method-generated datasets, we count how often a specific failure reason co-occurs with a given task feature, as shown in Fig.~\ref{fig:failure_taxonomy} (note that each task can have multiple feature labels but only one failure reason), so each cell in the heatmap shows the fraction of tasks that exhibit feature $i$ and elicit failure mode $j$, divided by the dataset size. \method-generated tasks cover more (task feature, failure mode) combinations than the original tasks (44 vs.39 out of 50 pairs). Broader coverage is desirable, since a robust test suite ideally elicits a variety of failure modes across different task features, rather than exploiting a single model vulnerability or a source of task difficulty. In the original task heatmap, 11 pairs are missing, whereas in the evolved heatmap, 6 pairs are missing. 

To further increase the diversity of the tasks (and elicited failures), one could turn this taxonomy into a reward bonus or into guidelines for the task Proposer, target each failure mode during the evolutionary search, and repeat generation for each seed task and desired failure combination. Despite not using this taxonomy a priori, \method already produces tasks that cover diverse failure categories and feature types, compared to the original benchmarks.

\section{Discussion \& Conclusion} As LLM capabilities advance, we need a scalable task generation approach to understand their current challenges and ways to improve them. We propose \method{}, a multi-agent evolutionary approach for generating \emph{solvable yet difficult} tasks. \method shows that hard tasks do not always require building new testing environments or APIs from scratch. Instead, existing tasks can be evolved and expanded by modifying instructions and recombining tool call sequences. 

\method has two clear advantages over existing data generation methods. (1) Evolving: Tasks evolve from the parent seed to target the solver’s competence, rather than converging on a one-size-fits-all static evaluation set. It also uses a gradient-free evolutionary search algorithm, which allows task generation to incorporate closed, frontier models for higher-quality generation. (2) Self-tuning: Unlike prior works that rely on humans for task design and refinement, \method is fully automated, with self-verification in a Dockerized environment and tuning based on the solver’s completion rate to ensure task reliability.

We apply \method to popular benchmarks, including AppWorld, $\tau$-bench, and AutomationBench, and evaluate the difficulty of the generated tasks relative to the original datasets. When the original tasks are too easy, \method yields tasks that are more difficult, leading to a broader range of solver failures. 

While we mainly use \method{} for evaluation, it can also be integrated into model training by scaling up training data generation with minimal human supervision. In particular, ~\citet{rutherford2024regretsinvestigatingimprovingregret} show that generating tasks with maximum outcome variance is ideal for learning, since they target the solver's learning frontier. We can first generate a new curriculum with \method{} and train the solver using reinforcement learning with verifiable rewards~\citep{lambert2025tulu3pushingfrontiers, Guo_2025} or apply filtered behavioral cloning to successful trajectories~\citep{hong2023zeroshotgoaldirecteddialoguerl}. This process of generating new tasks and training the solver can be repeated until the solver's learning plateaus, without collecting new human data or manually adjusting task difficulty. This can help close the model-improvement loop, from synthetic data generation for evaluation to training.

\section*{Limitations} \method uses gradient-free evolutionary search, so it can be applied to both open-weight and closed-weight models through API calls. However, the inference cost of proprietary models can be a limiting factor. We ran some ablations with different solver models (e.g., Gemini 3.6 Flash and Gemini 3.1 Flash Lite) on AppWorld using only a small subset of initial tasks. Our main objective was to show that \method applies to different solver types and is not overfitted to the Gemma solver, and to demonstrate different ways the generated tasks may be used. To make the datasets larger and more useful for future use, the initial task sets would need to be seeded with more parent tasks. 

\paragraph{Ethical Consideration.} As LLMs become more pervasive and integrated into our everyday lives and workflows, evaluating their failure modes is crucial for safe deployment and user control over AI systems. Our work contributes to safety by proposing new ways of generating task data to assess model capabilities, while minimizing  human supervision, reducing the engineering and data-generation burden. However, overreliance on AI systems based on their benchmark performance can be harmful, since benchmarks are designed to assess only certain aspects of a model's competence and can become saturated as models are continuously optimized against publicly released datasets.

\paragraph{AI Use Disclosure.} We used AI coding assistants to extend AlphaEvolve to different tool-use benchmarks and to debug implementations. We also used AI to polish the paper (both the writing and the formatting in Overleaf) and to polish the prompts used in our experiments. When LLMs are used as the evolutionary search orchestrator or as judges (e.g., evaluating task difficulty, labeling task features, and identifying failure reasons), their uses are described in the Methods section.



\bibliography{custom}

\clearpage

\appendix

\section{Appendix: Task Proposer Prompts}\label{appendix:task_proposer_prompts}
\begin{lstlisting}[language=Python, title=Task Proposer Prompt for AppWorld]
# We provide minimal guidance on how to mutate a task.

# Problem Context

Act as an expert software developer. Your task is to iteratively improve the provided codebase. The primary goal is to increase the scores on the provided evaluation metrics, where larger values are better.

Always adhere to best practices in Python coding.

## Task Generation Guidelines
When evolving the `get_task` function in `appworld_task_composer.py` for AppWorld:
- Return a Python dictionary containing the task structure.
- **IMPORTANT**: Make sure to use Python constants (e.g., `True`, `False`, `None`).
- Keep the dictionary within the `EVOLVE-BLOCK-START` and `EVOLVE-BLOCK-END`.
\end{lstlisting}

\begin{lstlisting}[language=Python, title=Task Proposer Prompt for $\tau$-bench banking]
# Problem Context

Act as an expert software developer. Your task is to iteratively improve the provided codebase. The primary goal is to increase the scores on the provided evaluation metrics, where larger values are better.

Always adhere to best practices in Python coding.

## Goal of Task Design: Uncovering Solver Blind Spots & Vulnerabilities
The overarching goal of task design is to **stress-test the solver agent to detect its blind spots, failure modes, and reasoning vulnerabilities**:
- **Isolate Problems & Failure Modes**: Design scenarios that isolate specific reasoning or tool-use difficulties (e.g., multi-constraint trade-offs, distractor options, subtle policy stipulations, ambiguous vs. precise user intents, strict eligibility prerequisites).
- **Creative Vulnerability Elicitation**: Be creative in how vulnerabilities are elicited through the conversational flow. For instance, craft realistic user personas that disclose constraints incrementally, present conflicting preferences that require careful KB verification, or test the solver's susceptibility to premature conclusions without proper verification.

## Task Generation Guidelines
When evolving the `get_task` function in `tau_banking_task_composer.py` for Tau Banking:
- Evolve the `task` Python dictionary defining the `user_scenario`, `initial_state`, `evaluation_criteria`, and `user_tools`.
- **IMPORTANT**: Make sure to use valid Python types (e.g., `True`, `False`, `None`) and follow the Pydantic schema for Tau2 Tasks. Do NOT use JSON types (e.g., `true`, `false`, `null`).
- Keep your changes strictly within the `EVOLVE-BLOCK-START` and `EVOLVE-BLOCK-END` comments.

## Creating New Tasks (Important)
Unlike standalone agents, Tau2 involves a conversational interaction. The solver agent starts completely blind and does NOT receive task instructions directly. Instead, the task instruction is conveyed to the solver dynamically through a **User Simulator**.
- To give the solver a new task, you must modify the `user_scenario.instructions` field to provide a new roleplay script, persona, and conversation flow for the User Simulator to act out. 
- You must also update the `evaluation_criteria.actions` to reflect the new expected sequence of ground truth API calls the solver should use to solve this new user scenario.
- **Do NOT modify or redefine `required_documents`**: The `required_documents` field is intentionally placed **outside** the `EVOLVE-BLOCK` at the end of `get_task` and contains the static Knowledge Base documents available for this task. You must NOT add, remove, or modify `required_documents` in your evolved code. Instead, use the provided documents outside the evolve block as fixed reference material to ground your `user_scenario` and `evaluation_criteria`.
\end{lstlisting}

\begin{lstlisting}[language=Python, title=Task Proposer Prompt for $\tau$-bench telecom]
# Problem Context

Act as an expert software developer. Your task is to iteratively improve the provided codebase. The primary goal is to increase the scores on the provided evaluation metrics, where larger values are better.

Always adhere to best practices in Python coding.

## Goal of Task Design: Uncovering Solver Blind Spots & Vulnerabilities
The overarching goal of task design is to **stress-test the solver agent to detect its blind spots, failure modes, and reasoning vulnerabilities**:
- **Isolate Problems & Failure Modes**: Design scenarios that isolate specific diagnostics or troubleshooting challenges (e.g., compounding network faults, multi-step dependency resolution, hidden device setting misconfigurations, edge-case roaming/data rules).
- **Creative Vulnerability Elicitation**: Be creative in how vulnerabilities are elicited through the conversational user simulation. For instance, craft personas with incomplete initial problem descriptions, misleading symptom descriptions that require rigorous systematic diagnosis, or step-dependent device toggles that punish guessing without checking.

## Task Generation Guidelines
When evolving the `get_task` function in the `tau_telecom_task_composer.py` file for Tau Telecom:
- Return a Python dictionary representing the new or modified Tau2 Task structure.
- **IMPORTANT**: Make sure to use valid Python types (e.g., `True`, `False`, `None`) and follow the Pydantic schema for Tau2 Tasks. Do NOT use JSON types (e.g., `true`, `false`, `null`).
- Keep the generated dictionary within the `EVOLVE-BLOCK-START` and `EVOLVE-BLOCK-END` comments.

## Creating New Tasks (Important)
Unlike standalone agents, Tau2 involves a conversational interaction. The solver agent starts completely blind and does NOT receive task instructions directly. Instead, the task instruction is conveyed to the solver dynamically through a **User Simulator**.
- To give the solver a new task, you must modify `user_scenario.instructions` (`reason_for_call`, `known_info`, `task_instructions`) and `user_scenario.persona` to provide the roleplay script and conversation flow for the User Simulator.
- You must set `initial_state.initialization_actions` to configure the initial device/network state (e.g. `set_network_mode_preference`, `break_vpn`, `turn_airplane_mode_on`, `set_user_location`, `turn_roaming_off`, `disable_roaming`, etc.).
- You must update `evaluation_criteria.actions` to reflect the expected sequence of ground truth API/device actions, and `evaluation_criteria.env_assertions` to verify that the issue is genuinely resolved (e.g. `assert_mobile_data_status`, `assert_internet_speed`, `assert_can_send_mms`, `assert_service_status`).
\end{lstlisting}

\begin{lstlisting}[language=Python, title=Task Proposer Prompt to AutomationBench]
# Problem Context

Act as an expert software developer and benchmark task designer. Your task is to iteratively improve the provided codebase. The primary goal is to increase the scores on the provided evaluation metrics, where larger values are better.

Always adhere to best practices in Python coding.

## Task Generation Guidelines
When evolving the `get_task` function in `automation_bench_task_composer.py` for AutomationBench:
- Return a Python dictionary containing the full task definition for AutomationBench.
- **IMPORTANT**: Make sure to use valid Python types (e.g., `True`, `False`, `None`, lists, dicts, strings, ints, floats). Do NOT use raw JSON literals (e.g. `true`, `false`, `null`).
- Keep the returned dictionary strictly within the `# EVOLVE-BLOCK-START` and `# EVOLVE-BLOCK-END` comments.
- Do NOT change the function signature `def get_task(task_name: str = None) -> dict:`.

## Task Structure & Required Fields
The dictionary returned by `get_task` must contain:
1. `example_id`: An integer ID (e.g., 4001, 4002, etc.).
2. `task_id`: A unique snake_case string identifier reflecting the domain and workflow (e.g., `"finance_disputed_invoice_escalation_4001"`, `"support_sla_breach_escalation_4002"`, `"sales_high_value_lead_routing_4003"`).
3. `domain`: The business domain (e.g., `"finance"`, `"support"`, `"sales"`, `"crm"`, `"devops"`, `"hr"`, `"marketing"`, `"operations"`).
4. `name`: A concise human-readable title describing the business workflow.
5. `description`: A dictionary containing:
   - `purpose`: High-level explanation of the workflow across multiple SaaS applications.
   - `difficulty_tier`: `"easy"`, `"medium"`, or `"hard"`.
   - `apps_used`: List of application names involved in the scenario (e.g., `["google_sheets", "gmail", "crm", "slack"]`).
6. `prompt`: The realistic user request / trigger instructions provided to the solver agent. It must contain the necessary context, identifiers, and business constraints without leaking the exact sequence of tool calls.
7. `initial_state`: The mock sandbox databases populated before evaluation starts (supports 47+ SaaS apps across categories):
   - **Google Workspace**:
     - `google_sheets`: `{"spreadsheets": {"<sheet_id>": {"title": "...", "rows": [{"invoice_id": "...", "client": "...", "amount": 123.45, "status": "...", "due_date": "...", "notes": "..."}, ...]}}}`
     - `gmail`: `{"inbox": [{"message_id": "...", "from": "...", "to": "...", "subject": "...", "body": "..."}, ...]}`
     - `google_drive`: `{"files": [{"file_id": "...", "name": "...", "mime_type": "...", "content": "..."}, ...]}`
     - `google_docs`: `{"documents": [{"document_id": "...", "title": "...", "body": "..."}, ...]}`
     - `google_calendar`: `{"events": [{"event_id": "...", "summary": "...", "start_time": "...", "end_time": "...", "attendees": [...]}, ...]}`
     - `google_workspace`: `{"users": [{"email": "...", "first_name": "...", "last_name": "...", "org_unit": "..."}, ...]}`
   - **CRM & Sales**:
     - `crm` / `hubspot` / `salesforce`: `{"accounts": {"<account_id>": {"account_name": "...", "status": "...", "primary_contact": "...", "assigned_ae": "...", "support_tier": "...", "tam_email": "...", "region": "...", "lifecycle_stage": "..."}, ...]}}`
   - **Communications**:
     - `slack`: `{"channels": {"<channel_id>": {"name": "finance-ops", "members": [...]}}, "messages": [{"channel": "...", "text": "..."}]}`
   - **Support & Helpdesk**:
     - `zendesk`: `{"tickets": {"<ticket_id>": {"account_name": "...", "subject": "...", "status": "...", "priority": "...", "description": "..."}, ...]}}`
     - `freshdesk`: `{"tickets": {"<ticket_id>": {"subject": "...", "description": "...", "email": "..."}, ...]}}`
   - **Project Management**:
     - `jira`: `{"issues": [{"issue_key": "...", "project": "...", "summary": "...", "description": "...", "priority": "...", "status": "..."}, ...]}`
     - `linear`: `{"issues": [{"id": "...", "title": "...", "team_id": "...", "description": "..."}, ...]}`
     - `asana`: `{"tasks": [{"task_id": "...", "name": "...", "projects": [...], "notes": "..."}, ...]}`
     - `notion`: `{"pages": [{"page_id": "...", "database_id": "...", "title": "...", "properties": {...}}, ...]}`
   - **HR & Marketing**:
     - `airtable`: `{"bases": {"<base_id>": {"<table_name>": [{"contact_id": "...", "name": "...", "email": "...", "company": "...", "minutes_attended": 45, "questions_asked": 2, "marketing_opt_in": True}, ...]}}}`
     - `bamboohr`: `{"employees": [{"id": "...", "name": "...", "status": "...", "role": "...", "department": "...", "manager": "...", "salary": "...", "home_address": "..."}, ...]}`
   - **Finance & Billing**:
     - `stripe`: `{"invoices": [{"invoice_id": "...", "customer": "...", "amount_due": 123.45, "status": "..."}, ...], "customers": [{"customer_id": "...", "email": "...", "name": "..."}, ...]}`
   - **Dev Platforms**:
     - `github`: `{"issues": [{"id": "...", "owner": "...", "repo": "...", "title": "...", "body": "..."}, ...], "invites": [{"email": "...", "role": "...", "team": "..."}, ...]}`
8. `info`:
   - `assertions`: List of target goal assertions evaluated after the solver execution:
     - `crm_account_field_equals`: parameters `{"account_id": "...", "field": "...", "expected_value": "..."}`
     - `gmail_message_sent_to_with_body_contains`: parameters `{"to": "...", "subject_contains": "...", "body_contains": ["...", "..."]}`
     - `slack_message_posted_to_channel`: parameters `{"channel": "...", "text_contains": ["...", "..."]}`
     - `google_sheets_row_exists`: parameters `{"spreadsheet_id": "...", "cell_contains": {"<col>": "<val>"}}`
     - `zendesk_ticket_field_equals`: parameters `{"ticket_id": "...", "field": "status"|"priority", "expected_value": "..."}`
     - `jira_issue_exists_in_project`: parameters `{"project": "...", "summary_contains": "..."}`
     - `hubspot_contact_has_tag`: parameters `{"email": "...", "expected_tag": "..."}`
     - `google_workspace_user_exists`: parameters `{"primary_email": "..."}`
     - `github_invite_sent`: parameters `{"email": "...", "expected_role": "..."}`
   - `guardrails`: Negative policy constraints that must NOT be triggered:
     - `slack_message_not_posted_to_channel`: parameters `{"channel": "general"}`
     - `gmail_message_not_sent_to`: parameters `{"to": "unauthorized@example.com"}`
     - `gmail_message_body_not_contains`: parameters `{"to": "...", "forbidden_patterns": ["TAX-ID-", "SSN-", ...], ...}`
     - `github_invite_role_not_equals`: parameters `{"email": "...", "forbidden_role": "admin"}`
     - `slack_message_text_not_contains`: parameters `{"channel": "...", "forbidden_patterns": [...]}`
9. `solution`: The golden ground truth sequence of actions `[{"tool": "<tool_name>", "arguments": {...}}, ...]`. Supported tool names include:
   - `google_sheets_get_rows` (`spreadsheet_id`, `query`)
   - `google_sheets_append_row` (`spreadsheet_id`, `row`)
   - `crm_get_account` (`account_id`)
   - `crm_update_account` (`account_id`, `fields`)
   - `gmail_send_message` (`to`, `subject`, `body`)
   - `slack_post_message` (`channel`, `text`)
   - `zendesk_update_ticket` (`ticket_id`, `fields`)
   - `jira_create_issue` (`project`, `summary`, `description`, `priority`)
   - `hubspot_update_contact` (`email`, `tags`)
   - `airtable_get_records` (`base_id`, `table`)
   - `bamboohr_get_employee` (`name`)
   - `google_workspace_create_user` (`primary_email`, `name`, `org_unit_path`)
   - `github_invite_member` (`email`, `role`, `team_slugs`)

## Crucial Quality & Learnability Requirements
- **100% Ground Truth Solvability**: The `solution` MUST perfectly satisfy all `assertions` and violate 0 `guardrails` when executed sequentially in the sandbox. Any task failing ground-truth verification receives 0 reward.
- **Multi-App Composition**: Create complex, realistic 3 to 6 step workflows across at least 3 distinct SaaS tools (e.g. read Sheets -> update CRM -> send tailored Gmail -> post Slack alert).
- **Curriculum Difficulty & Learnability**:
  - The evaluation metric optimizes for tasks with an intermediate solver completion rate (~0.3 to 0.7) and high trial variance, as these provide maximum learnability signal.
  - Avoid overly trivial tasks (100% completion by solver) or unsolvable/confusing instructions (0% completion).
  - Use subtle business constraints, ambiguous entity lookups that require checking database rows, or strict formatting guardrails to challenge the solver model.
- **Novelty & Diversity**:
  - Generate tasks across diverse domains (Sales Ops, HR onboarding, Incident Management, Billing Reconciliations, Security Access).
  - Explore novel failure reasons and corner cases different from past attempts.
\end{lstlisting}

\begin{figure*}[htbp]
\centering
\begin{tcolorbox}[colback=gray!5, colframe=black, boxrule=1pt, arc=0pt, width=\textwidth]
Instruction: \{instruction\} \\
\\
Expected Ground-Truth Final States: \{expected\_state\_changes\} \\ \\

Ground-Truth Tool Calls: \{ground\_truth\_calls\}\\
\\
Failure Reason Summary: \{failure\_reason\}\\ \\

Failure Trajectories: \{failure\_trajectories\} \\ \\

Task: You are an expert LLM judge. Evaluate if the failure described in the 'Failure Reason' is a trivial failure by the solver. \\
A trivial failure is one where the solver basically did the right thing but failed due to a minor string matching detail, overly strict evaluation, or a harmless deviation. \\ \\ 

Score it as 1 (Trivial Failure) or 0 (Not a Trivial Failure). \\ \\ 

Example of a Trivial Failure (Score 1): \\
"The agent failed because it set the refund transaction description to just the original purpose ('movies') instead of combining it into a full description like 'Refund for movies payment request' as expected." \\ 
(This should be leniently accepted since the solver included 'movies', so it's a trivial failure). \\ \\ 

Example of a Trivial Failure (Score 1): \\ 
"The agent failed because it used an incorrect description ('Refund for movies') for the returned Venmo payment instead of properly describing it as returning the accidental payment." \\ 
(This should be tolerated and evaluated as a trivial failure). \\ \\ 

Example of a Non-Trivial Failure (Score 0): \\ 
"The task required downloading 5 songs, and the agent only downloaded 4 songs." \\
(This is a material failure, not a trivial one). \\ \\ 

Output your response in JSON format with the following keys: \\
- "score": An integer (0 or 1). \\
- "justification": Exactly 1 concise sentence explaining the score. \\ \\ 

Response format: \\
\{ \\
  "score": ..., \\ 
  "justification": "..." \\ 
\}
\end{tcolorbox}
\caption{LLM-judge prompt to remove tasks with trivial failures caused by instruction ambiguity.}
\label{appendix:trivial_solver_failure}
\end{figure*} 

\clearpage
\section{Appendix: Datasets}\label{appendix:dataset}
\noindent \textbf{AppWorld}~\citep{trivedi2024appworldcontrollableworldapps} includes a set of 4,571 executable APIs covering everyday mobile applications like Spotify, Amazon, Venmo, and Gmail. Tasks look like: `Send a Venmo request to \{a friend's name\}" and ``Download a song by \{artist name\} on Spotify and add it to my playlist", and require the LLM agent to take actions on behalf of the assigned user (i.e., human supervisor). The dataset is split into a larger test set and relatively smaller datasets for training and validation, which include the initial database states, task specifications, and ground truth API sequences. 

\textbf{Our design choice.} To verify whether a task is fully solved, we initialize two identical environments with a given database state, execute the ground-truth API calls and the solver's actions in parallel across each copy, and compare their final states. For any key-value pairs in the final state dictionary that do not match between the ground truth and the solver, we run an LLM judge as a fallback state-equivalence check to filter out differences caused by formatting or exact string mismatches.

\noindent \textbf{$\tau$-bench} is a tool use benchmark for evaluating the model's capability as a customer-facing service agent across different domains. In particular, we use telecom~\citep{barres2025tau2benchevaluatingconversationalagents} and banking~\citep{shi2026tauknowledgeevaluatingconversationalagents}. In both domains, since the instruction is provided through the interactions with the user simulator, the user ends the conversation when their goal is satisfied.

\noindent \textbf{$\tau$- telecom}~\citep{barres2025tau2benchevaluatingconversationalagents} evaluates a dual-control, multi-turn dialogue setup, where both the user and the agent are capable of making non-overlapping sets of tool calls. The LLM solver is prompted to act as a customer service agent resolving the simulated user's problem. Certain tool calls can only be executed by the user, so the agent must guide the user through step-by-step instructions to achieve the desired tool execution. In our setup, we use the same base model for both roles. The user is given an additional prompt specifying their persona and guidelines on how to respond to requests or information-gathering questions made by the agent. The full dataset features 2,285 tasks across three user persona types alongside different tasks. When evaluating  solver attempts against the ground-truth tool calls, we need to check for all environment changes, including those invoked by the solver (agent) as well as the user. 

\textbf{Our design choice.} We allow the solver to continue interacting with the user for maximum 25 turns per task (since the longest task requires 12 actions), and three consecutive errors in a conversation result in early terminations with a 0 reward. 
Following ~\citet{barres2025tau2benchevaluatingconversationalagents}, we only allow atomic solver actions per turn.

\noindent \textbf{$\tau$- Banking}~\citep{shi2026tauknowledgeevaluatingconversationalagents} is specialized for testing an agent's document retrieval and long-context reasoning capabilities. The agent must navigate large, unstructured knowledge bases covering financial topics, like checking and savings accounts, credit card terms, annual percentage yields, and internal protocols to resolve users' transaction disputes and product inquiries. The internal database consists of 698 documents spanning 21 product categories, and the dataset has 97 complex problems that require information retrieval, multi-step tool execution, and policy constraint resolution, while navigating multi-turn user dialogue. Each task requires on average 18.6 documents and 9.5 tool calls, with 33 maximum number of tool calls. Different configurations are available for retrieval: dense retrieval with text embeddings, sparse retrieval with BM25, terminal use (knowledge base is exported as files and the agent can use shell commands), and golden (ground truth documents are directly supplied in context, no retrieval required). 

\textbf{Our design choice.} Our solvers are evaluated with BM25 and golden retrieval, and allowed up to 50 max turns and 3 consecutive errors per task. Following ~\citet{shi2026tauknowledgeevaluatingconversationalagents}, we allow solvers to execute simultaneous tool-calls in a single turn.

Our evaluation covers both state-mutating and non-mutating tasks. For state-mutating tasks, we initialize the environment and compare the resulting state changes between the ground truth execution and the solver's tool calls. We assign a score of 1 for an exact match and 0 otherwise. For non-mutating tasks (e.g., escalating an issue to a human agent), we evaluate action matching based on whether the agent makes the correct tool with the correct parameters. 

 \noindent \textbf{AutomationBench}~\citep{shepard2026automationbench} is a complex domain built on 47 simulated SaaS applications, such as Gmail, Google Sheets, Slack, Salesforce, Zendesk, Jira, and HubSpot. The dataset consists of 657 tasks spanning Finance, HR, Marketing, Operations, Sales, and Support with a max turn of 50 steps. The agent is tested on automating SaaS workflows while navigating business objectives and guardrails. The agent is also tested on autonomous API discovery, where it needs to figure out which apps are relevant and which REST API endpoints to execute. Unlike the other domains, evaluations with AutomationBench specify guardrails and target changes to the final states after tool-call execution. The solver's goal is to match all target changes executed by the ground truth model and avoid any changes to the guardrails (violations nullify the successful changes). 

 \textbf{Our design choice.} To avoid reward sparsity on challenging tasks, we allow partial rewards on AutomationBench.

\begin{figure*}[htbp]
\centering
\begin{tcolorbox}[colback=gray!5, colframe=black, boxrule=1pt, arc=0pt, width=\textwidth]
\textbf{AppWorld} \\ 
\# EVOLVE-BLOCK-START \\
return \{
	db\_changes: \{ initial user database states \}, \\
	full\_description: user account information (e.g., access tokens, emails, tool-discovery instructions to the solver),  \\
    instruction: task-specific instruction to the solver, \\ 
    ground\_truth: summary of ground-truth tool calls, \\ 
    ground\_truth\_calls: list of ground-truth tool calls (including function parameters), \\ 
\} \\
\# EVOLVE-BLOCK-END \\
\\
\textbf{$\tau$-bench (telecom)} \\ 
\# EVOLVE-BLOCK-START \\ 
return \{ \\
	user\_scenario: { persona: ..., instruction: .... }, \\
	initial\_state: .., \\
	evaluation\_criteria: \{ \\ 
		actions: [name: tool\_call\_x, arguments: \\\{params\_to\_x\} ], \\ 
		communicate\_info: [], \\
		reward\_basis: [DB / action],  \\
\# DB means stateful evals, actions are read-only tasks
	\}, \\
	user\_tools: [tools available to the user]
\} \\
\# EVOLVE-BLOCK-END \\ 

\textbf{$\tau$-bench (banking)} \\ 
\# EVOLVE-BLOCK-START \\
return \{ \\
	user\_scenario: \{ persona: ..., instruction: .... \},\\
	initial\_state: ..,\\
	required\_documents: [\{id: .., content: ... \}, ], \\ \# solver (agent) gets access to all 600+ documents for knowledge retrieval through BM search, but the task generator (AlphaEvolve) only gets the relevant documents directly as context (and the proposer is not allowed to change the document content) \\
	evaluation\_criteria: \{ \\
		actions: [name: tool\_call\_x, arguments: \{params\_to\_x\} ], \\
		communicate\_info: [], \\
		reward\_basis: [DB / action], \\
\# DB means stateful evals, actions are read-only tasks
	\},\\
	user\_tools: [tools available to the user]\\
\} \\
\# EVOLVE-BLOCK-END \\
\end{tcolorbox}
\caption{Task template for AppWorld and $\tau$-bench.}
\label{appendix:task_template}
\end{figure*}

\begin{figure*}[htbp]
\centering
\begin{tcolorbox}[colback=gray!5, colframe=black, boxrule=1pt, arc=0pt, width=\textwidth]
\textbf{AutomationBench}\\ \# EVOLVE-BLOCK-START \\
return \{
	domain: finance, \\ 
	name: \# 1 sentence task summary, \\
	description: \{ purpose: ...,  \\difficulty\_tier: ..., apps\_used: [] \}, \\
	prompt: \# instruction to the solver, \\
	initial\_state: \{
		\# initial states for relevant apps  \\(e.g., google\_sheets, gmail, slack)\}, \\
	info: \{ \# specifies evaluation criteria  \\
		assertions: [ \# list of expected state changes ], \\
		guardrails: [ \# list of unwanted state changes ] \\
	\}, \\
	solution: [ \# list of ground truth tool calls ] \\
\} \\ 
\# EVOLVE-BLOCK-END
\end{tcolorbox}
\caption{Task template for AutomationBench.}
\end{figure*}


\begin{figure*}[htbp]
\centering
\begin{tcolorbox}[colback=gray!5, colframe=black, boxrule=1pt, arc=0pt, width=\textwidth]
\small 
You are an expert evaluator assessing the complexity of tool-calling tasks. Your goal is to evaluate a given task scenario and output its difficulty level on a scale of 1 to 3 (1: Easy, 2: Medium, 3: Hard).\\ \\

\#\#\# Difficulty Guidelines \\ 
The level is assigned based on the notion of difficulty based on the extent of planning, reasoning, and interaction requirements and engineered hurdles and distractors. \\ \\ 

When evaluating interaction requirements, consider the difference between weak and strong interactions: \\
- Weak Interaction Requirement: The task could theoretically be solved in one shot if the agent had perfect knowledge of the APIs (e.g., direct filtering and passing IDs). However, in practice, agents still benefit from incremental exploration and trial-and-error. \\ 
- Strong Interaction Requirement: The task cannot be solved in one shot, even theoretically. The agent must interact with the environment to retrieve missing context (e.g., reading free-form text from a note or email), use language understanding to parse that information, and then use those insights to plan subsequent API calls. Harder tasks typically feature these strong interaction requirements.\\ \\ 

\#\#\# In-Context Examples \\ \\

Easy (Level 1) \\ \\
- How many Venmo friends have I made since the start of \{February\} \{last\} year? \\ 
- Place an order for all \{t-shirts\} in my Amazon wish list \\ \\

Medium (Level 2) \\
- Add a new habit tracking log note for today in my Simple Note account. It should be the same as yesterday, except I did not work out today \\ 
- My manager has asked me for my headshot picture over email, to which I have drafted a response. I mistakenly attached my \{resume\} to it instead. Please change it to the headshot from my file system, and send the email. \\ \\

Hard (Level 3) \\ 
- I want to transfer my music library to Spotify. I’ve exported the data from my previous software to a text file located at \texttt{\{"\textasciitilde/backups"\}}. Please make a Spotify account for me and import my song, album, and playlist libraries into it as per that file. \\
- I’m job hunting and have prepared a cover letter template, a list of potential employers, and their email addresses in my Simple Note account. My \{resume\} is stored at ``\{\textasciitilde/documents/work/\}''. Please send individual emails to these employers with the subject ``\{Job Application Submission\}", using the cover letter template for the body, and attach my \{resume\}. \\ \\

\#\#\# Task to Evaluate \\ 
Task Scenario: \{task\_scenario\} \\ \\

\#\#\# Instructions \\
1. Briefly reason about the extent of planning, reasoning, and interaction requirements (weak vs. strong) needed to solve the task in exactly 1 sentence. \\
2. Output a valid JSON dictionary with two keys: "reasoning" (string) and "score" (integer, 1-3). 

\end{tcolorbox}
\caption{\textbf{LLM-judge rating task difficulty.} AppWorld Benchmark~\citep{trivedi2024appworldcontrollableworldapps} tasks are scored on a scale of 1–3 (easy to hard) by human annotators based on the amount of planning, reasoning, and interaction required to solve the task, as well as the presence of hurdles and distractors. We included these examples (two instructions for each level from Table 10 of their paper) as in-context examples in the LLM-judge prompt. We also evaluated the reliability of Gemini 3.6 Flash as a judge by checking its accuracy on the 732 human-labeled AppWorld tasks. Gemini 3.6 Flash achieved: overall accuracy of 58.74\%, 34.9\% on easy tasks, 64.34\% on medium tasks, and 70.2\% on hard tasks. }
\label{appendix:llm_difficulty_prompt}
\end{figure*}

\clearpage

\begin{figure*}[t]
    \centering
    \includegraphics[width=\textwidth]{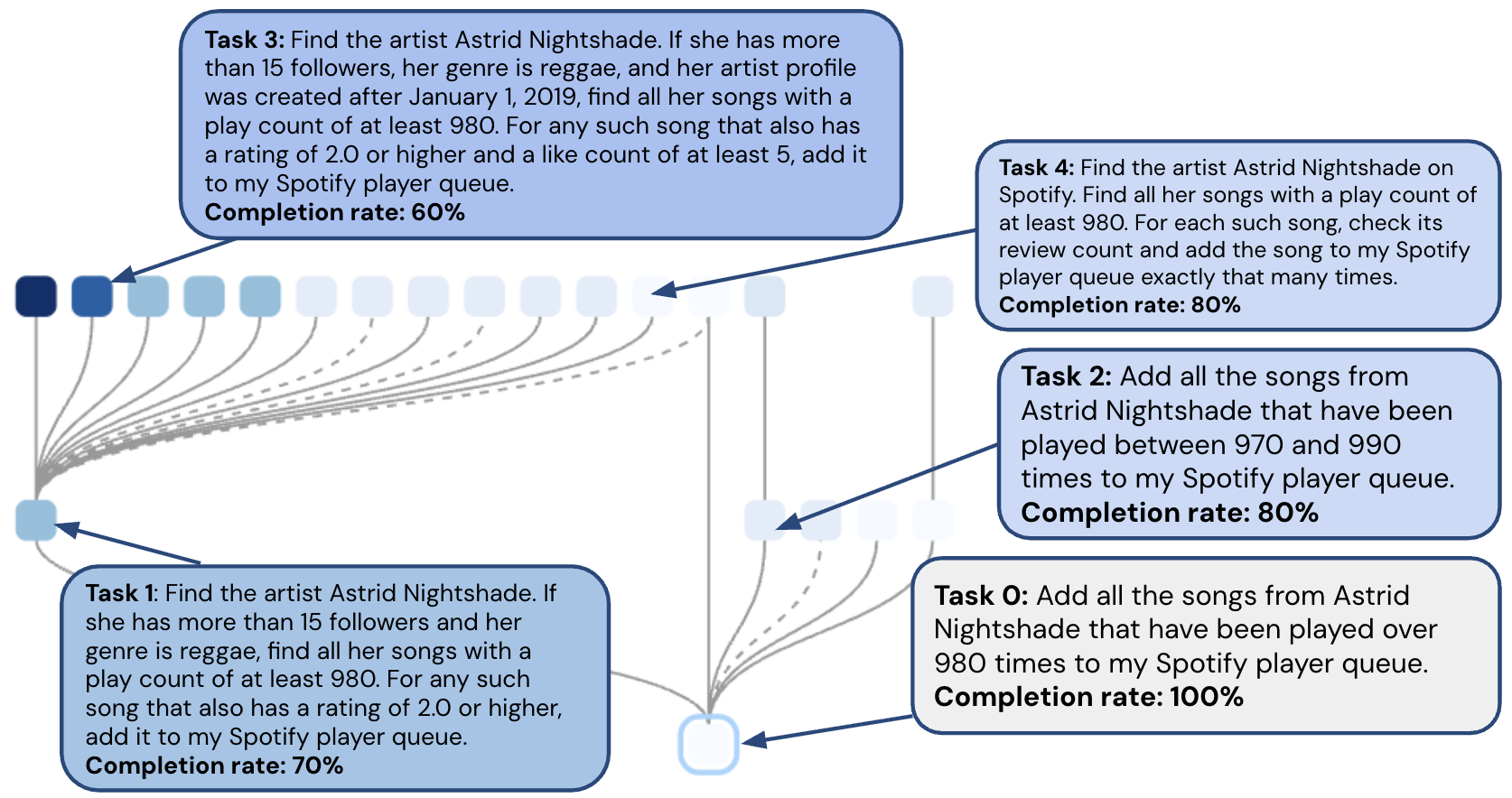}
    \caption{AppWorld task evolved using Gemini 3.1 Flash Lite.}
    \label{fig:task_lineage1}
\end{figure*}

\begin{figure*}[t]
    \centering
    \includegraphics[width=\textwidth]{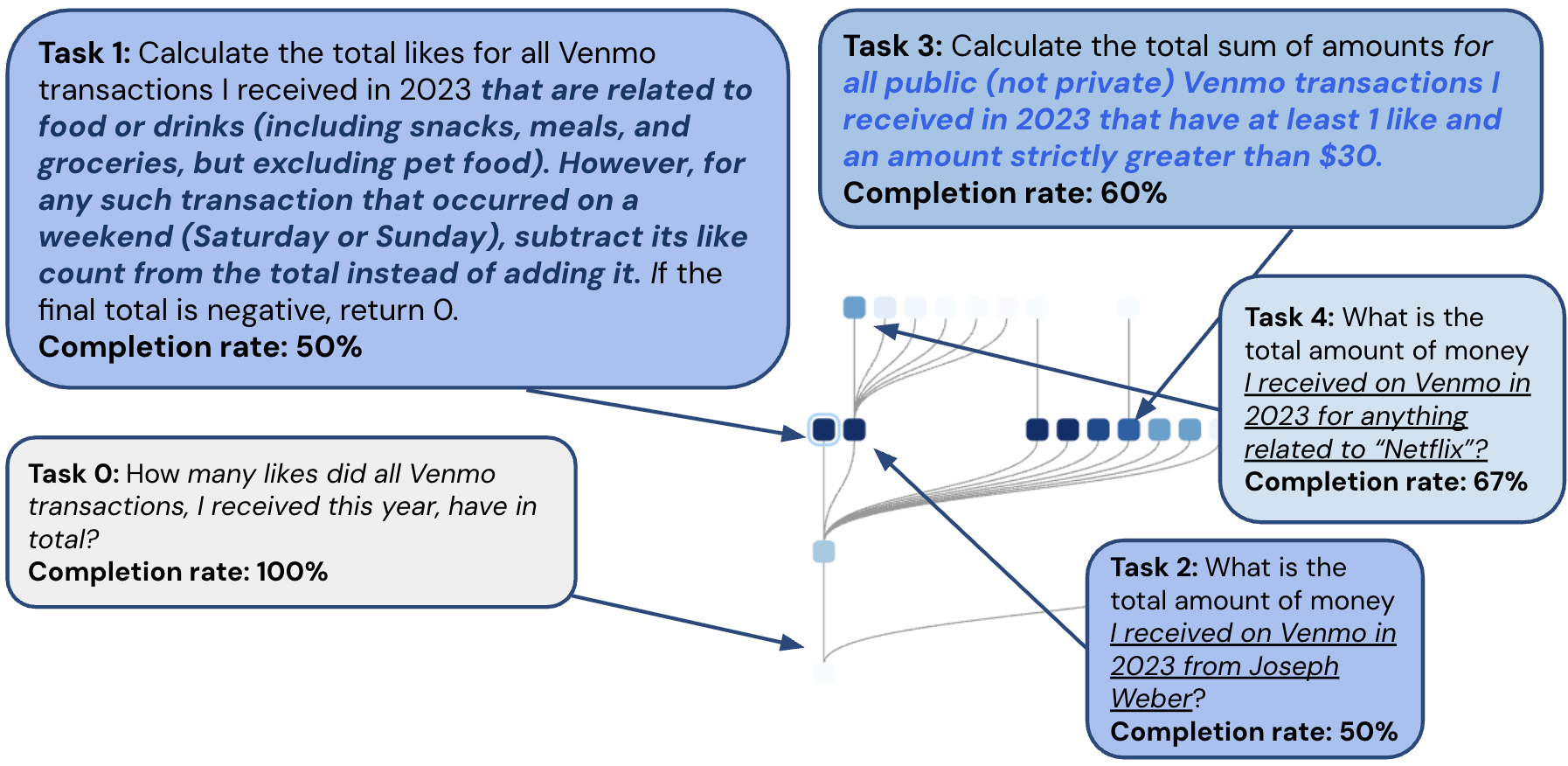}
    \caption{AppWorld task evolved using Gemini 3.6 Flash.}
    \label{fig:task_lineage2}
\end{figure*}

\begin{figure*}[t]
    \centering
    \includegraphics[width=\textwidth]{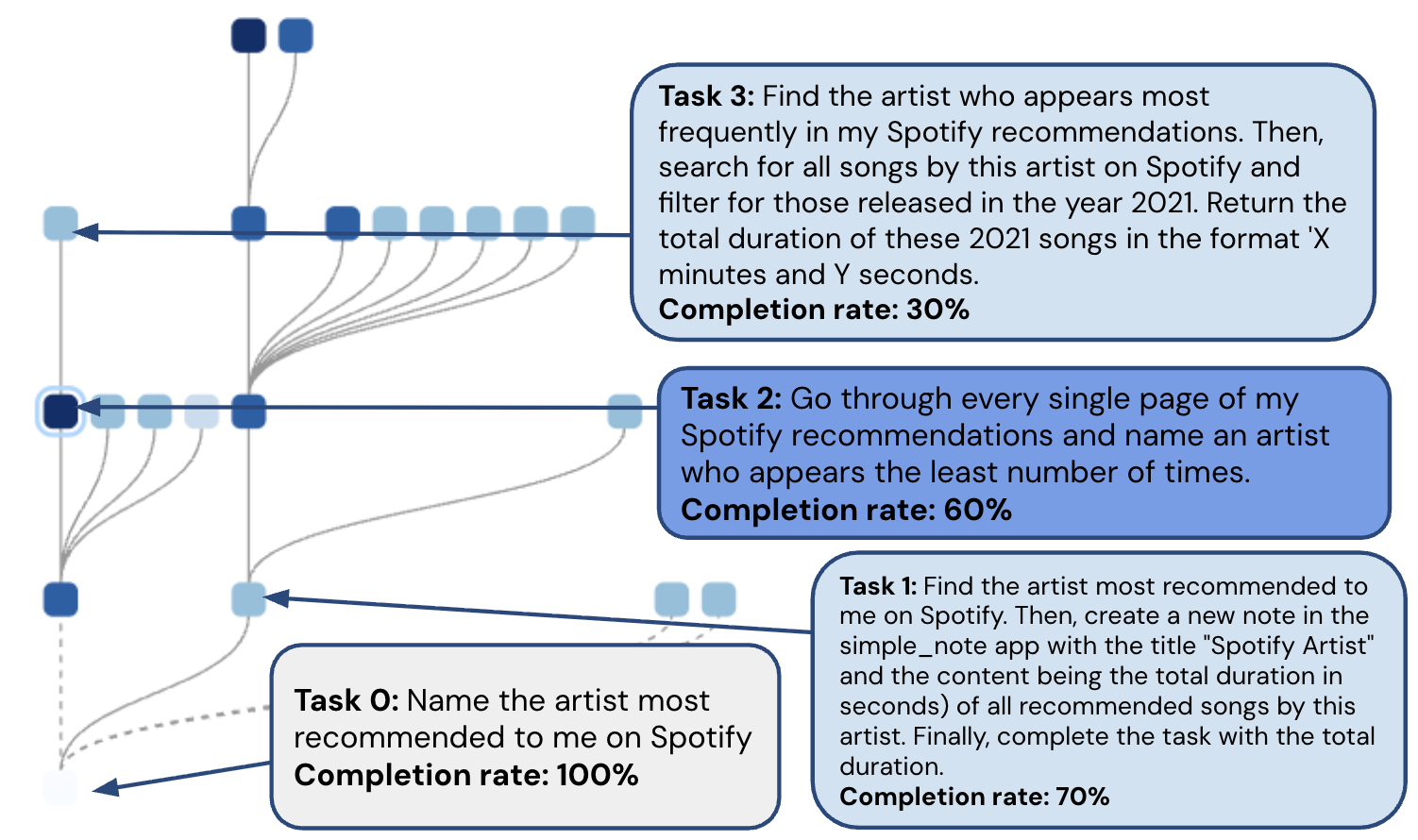}
    \caption{AppWorld task evolved using Gemma 4 31B-it.}
    \label{fig:task_lineage}
\end{figure*}

\begin{figure*}[t]
    \centering
    \begin{subfigure}[t]{0.48\linewidth}
        \centering
        \includegraphics[width=\linewidth]{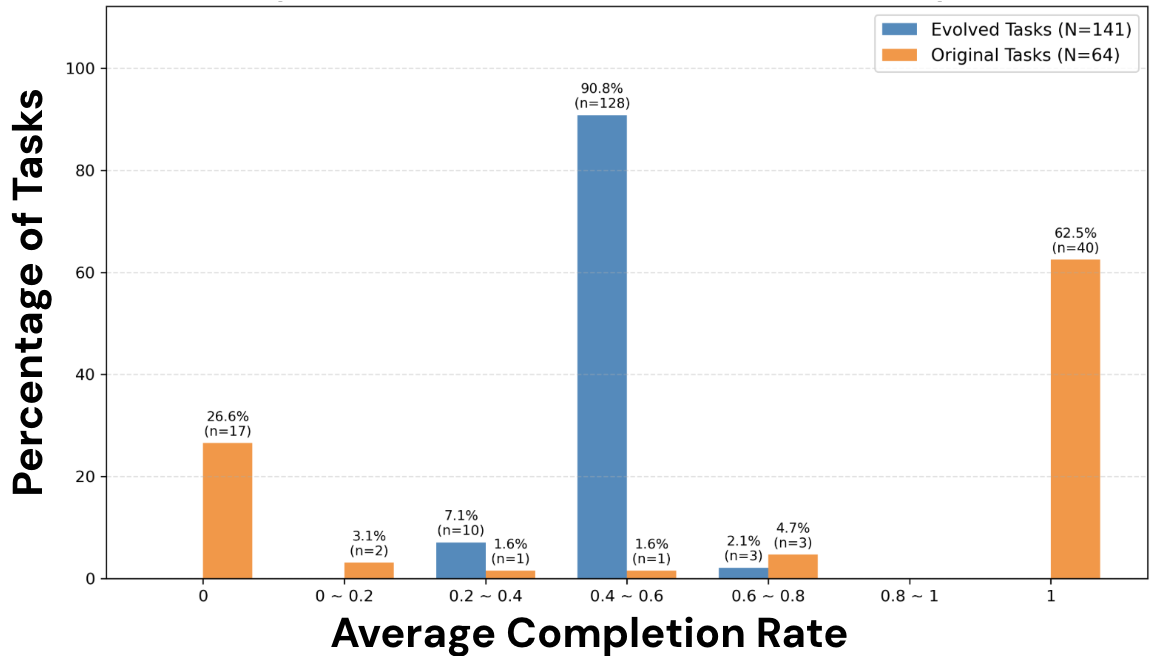}
        \caption{Average completion rate comparison of initial ($n=64$) and \method ($n=141$) tasks using Gemini 3.1 Flash Lite.}
    \end{subfigure}
    \hfill
    \begin{subfigure}[t]{0.48\linewidth}
        \centering
        \includegraphics[width=\linewidth]{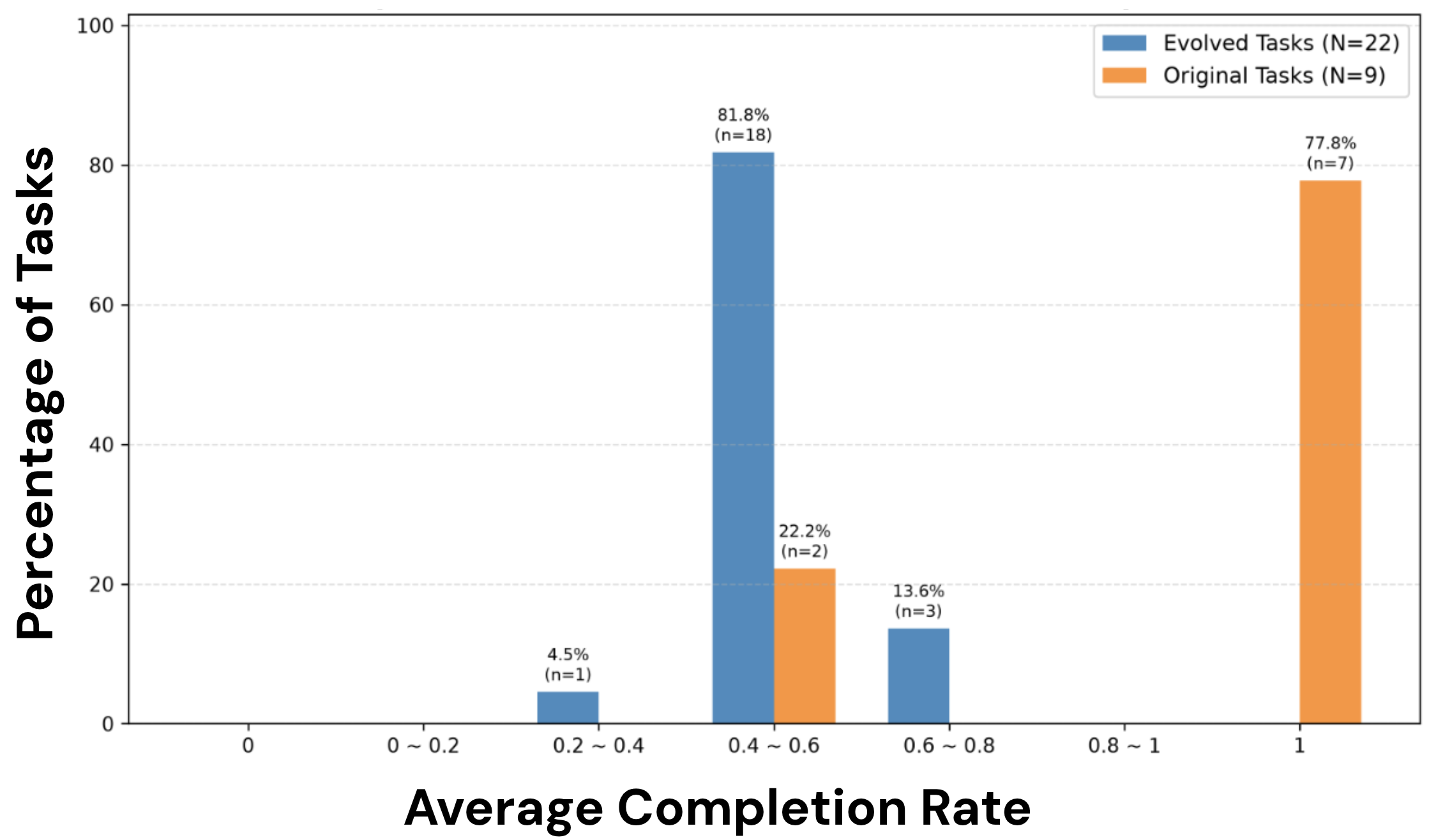}
        \caption{Average completion rate comparison of initial ($n=9$) and \method ($n=22$) tasks using Gemini 3.6 Flash.}
    \end{subfigure}

    \caption{Applications of \method{} to AppWorld with different Gemini models as target solvers.}
    \label{appendix:appworld_completion_rates}
\end{figure*}

\begin{table*}[htbp]
    \centering
    \small
    \setlength{\tabcolsep}{4pt}
    \caption{\textbf{Taxonomy of desirable task features.} We used Gemini 3.6 Flash as an LLM judge to label whether each of the specified features is present in a given task instruction. This taxonomy was developed from human review of tasks generated with AppWorld and from error analyses provided by the following benchmarks and papers: AppWorld~\citep{trivedi2024appworldcontrollableworldapps}, AutomationBench~\citep{shepard2026automationbench}, $\tau$-banking~\citep{shi2026tauknowledgeevaluatingconversationalagents}, SWEBench~\citep{jimenez2024swebenchlanguagemodelsresolve}, GDPval~\citep{patwardhan2025gdpvalevaluatingaimodel}, and FactoryBench~\citep{merzouki2026factorybenchevaluatingindustrialmachine}.}\label{appendix:task_feature_taxonomy}
    \begin{tabular}{|p{0.16\linewidth}|p{0.14\linewidth}|p{0.28\linewidth}|p{0.34\linewidth}|}
        \hline
        \textbf{Task property} & \textbf{Source Dataset} & \textbf{Description} & \textbf{Examples} \\ \hline
        Specification & AppWorld & Explicitly requires the output to be formatted in an exact way that goes beyond just providing the content. & Title files as \texttt{date\_owner's last name\_creation date}; give comma-separated answers  or write as \texttt{answer: \{value\}} (if specific formatting requirements are made in the instruction, the evaluator must check for instruction following accuracy) \\  \hline
        Reasoning & AppWorld & Requires an additional reasoning or transformation step that is not natively handled by an API call. & Alphabetically sorting the output list; retrieving only the last four digits of a credit card; mathematical calculations. \\ \hline
        Distractor & AppWorld, AutomationBench, SWEBench & Includes distractors or irrelevant information that are unnecessary for the main tool calls, requiring the agent to filter out the noise. & Solver needs to retrieve the most recent order, but the API returns a list containing the past $N$ items which include distractor names; adding irrelevant details or variable/ entity names that are unused in the solution. \\  \hline
        Intermediate tool-call & AutomationBench, $\tau$-banking & Does not explicitly state which API should be used, or it references entities where the necessary information to call the API must first be retrieved through intermediate discovery. & "Return my clothes", but the return policy document must be discovered; "Email my roommates", but the roommates' identities and their contact information are unknown in the prompt and need to be discovered through additional API calls. \\  \hline
        Constraint \& Filter & $\tau$-banking & Includes strict logical constraints, conditional (if-then) logic, or overriding priorities and exceptions. & Constraints: "Only download songs by the artist with N followers"; Conditionals: "If I have more than 5 friends, send them a text individually; otherwise send them a group message."; Overriding Priorities/Exceptions: e.g., A general rule exists (return items without expiration date), but a specific overriding policy applies (clothes expire after N days), and there is an exception to that override (except for certain brands). \\ \hline
    \end{tabular}
\end{table*}

\begin{figure*}[t]
    \centering
    \begin{subfigure}[t]{0.48\linewidth}
        \centering
        \includegraphics[width=\linewidth]{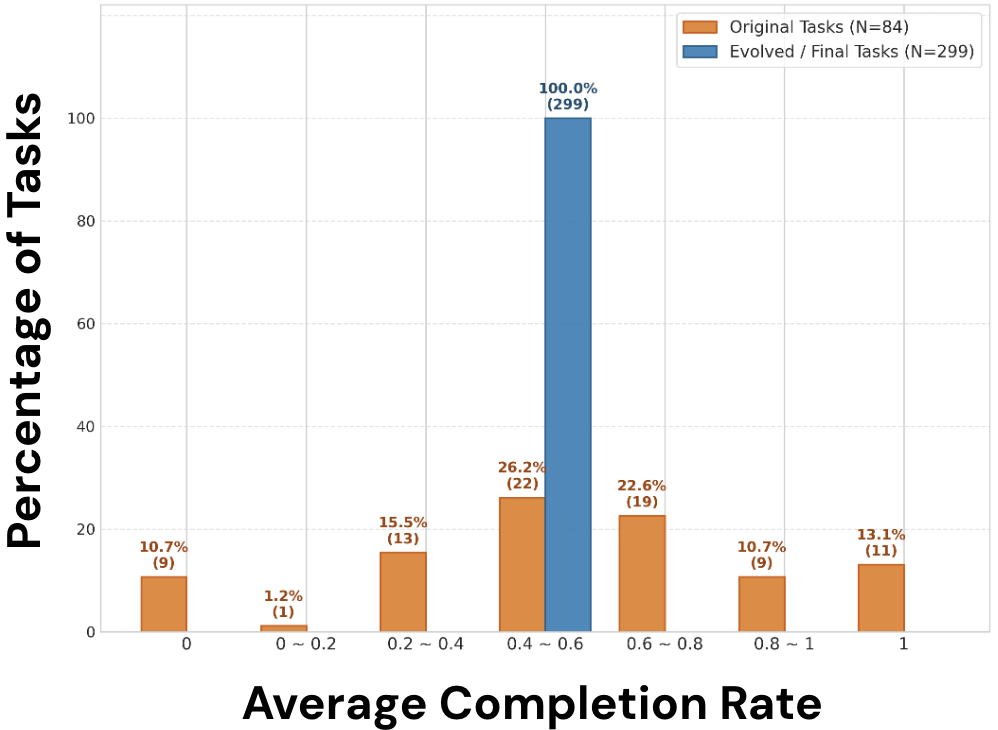}
        \caption{Average completion rate comparison of initial ($n=84$) and evolved ($n=299$) tasks in $\tau$-telecom.}
    \end{subfigure}
    \hfill
    \begin{subfigure}[t]{0.48\linewidth}
        \centering
        \includegraphics[width=\linewidth]{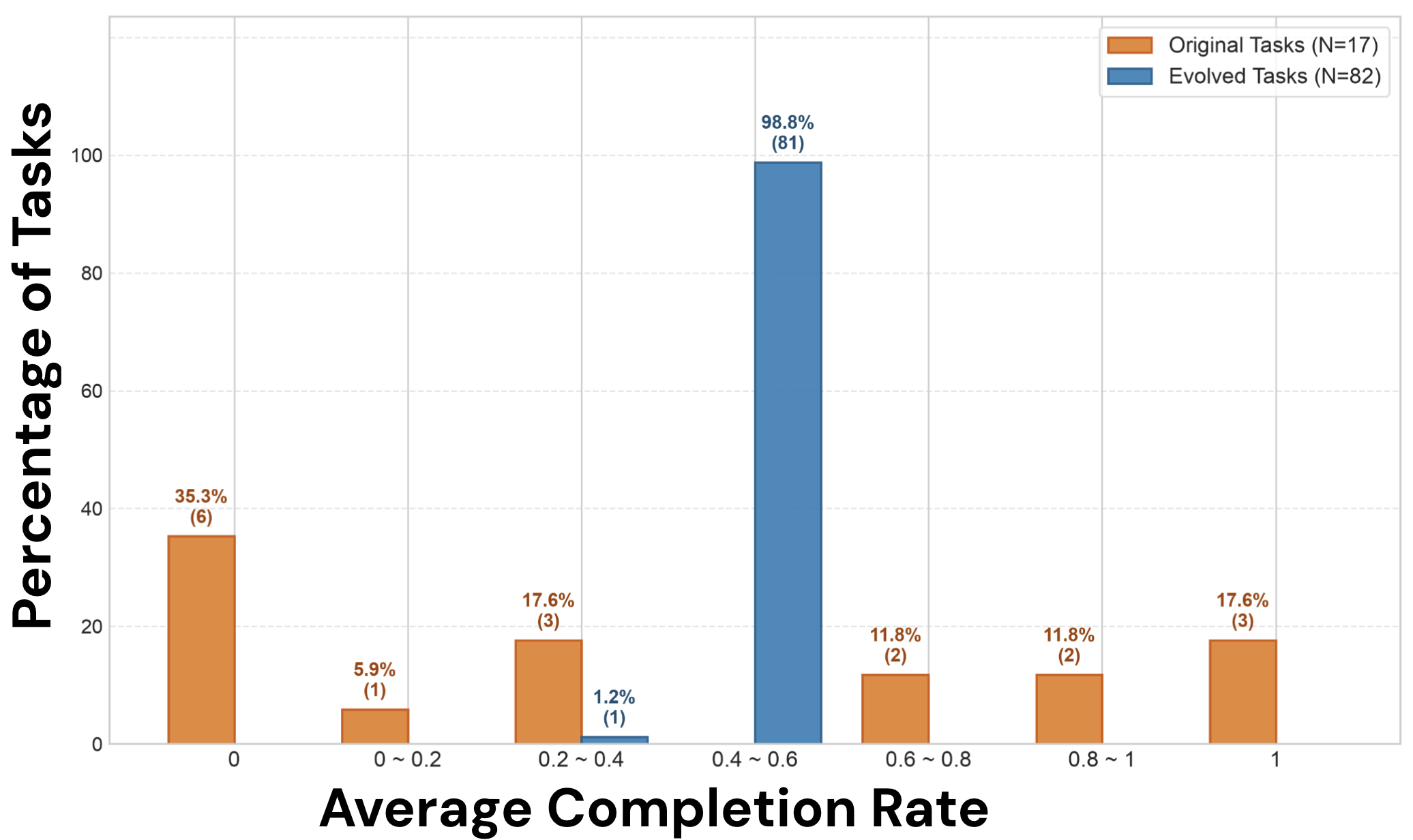}
        \caption{Average completion rate comparison of initial ($n=17$) and evolved ($n=82$) tasks in $\tau$-banking (BM25 retrieval).}
    \end{subfigure}

    \vspace{1em}

    \begin{subfigure}[t]{0.48\linewidth}
        \centering
        \includegraphics[width=\linewidth]{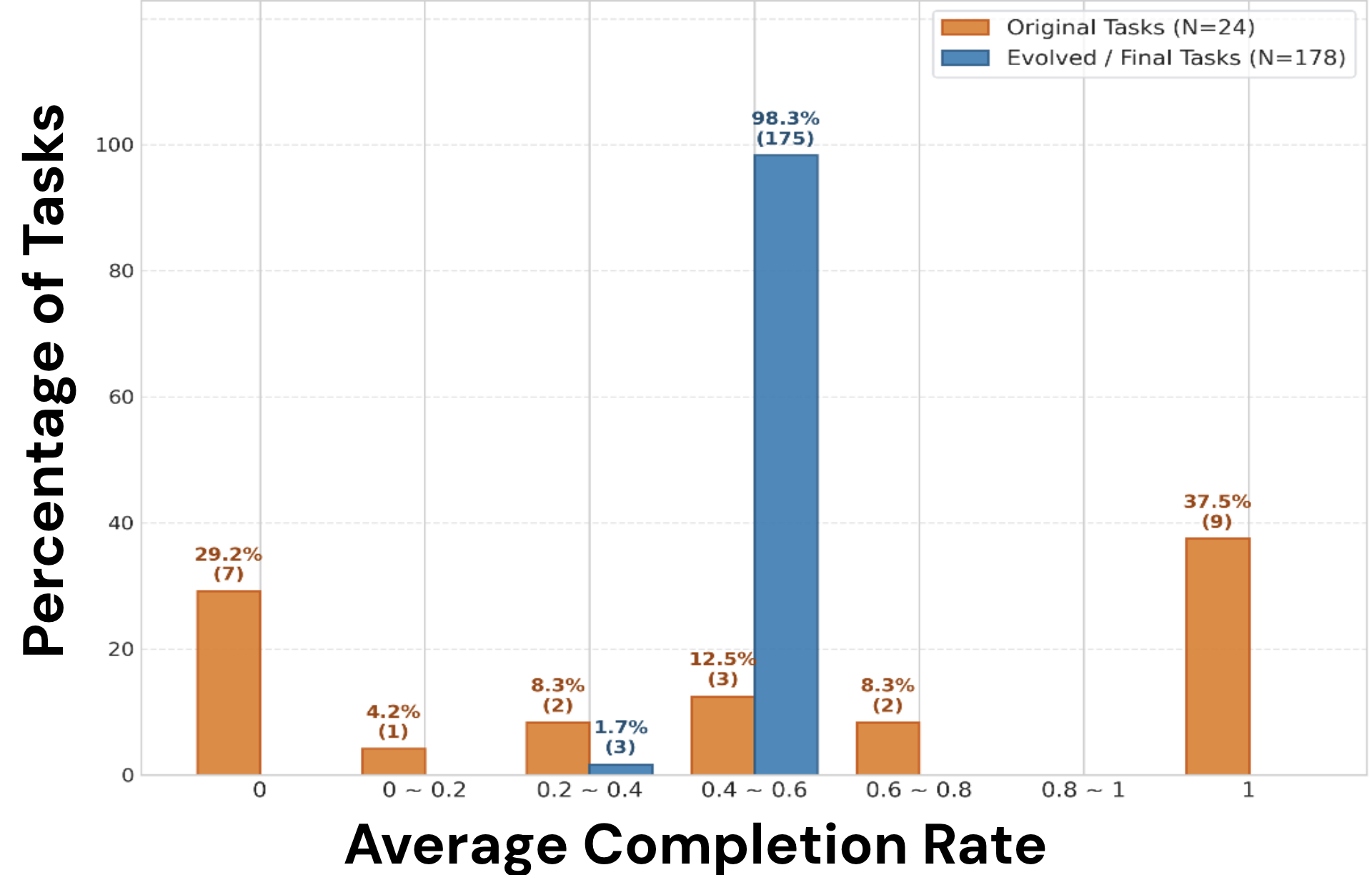}
        \caption{Average completion rate comparison of initial ($n=24$) and evolved ($n=178$) tasks in $\tau$-banking (golden retrieval).}
    \end{subfigure}
    \hfill
    \begin{subfigure}[t]{0.48\linewidth}
        \centering
        \includegraphics[width=\linewidth]{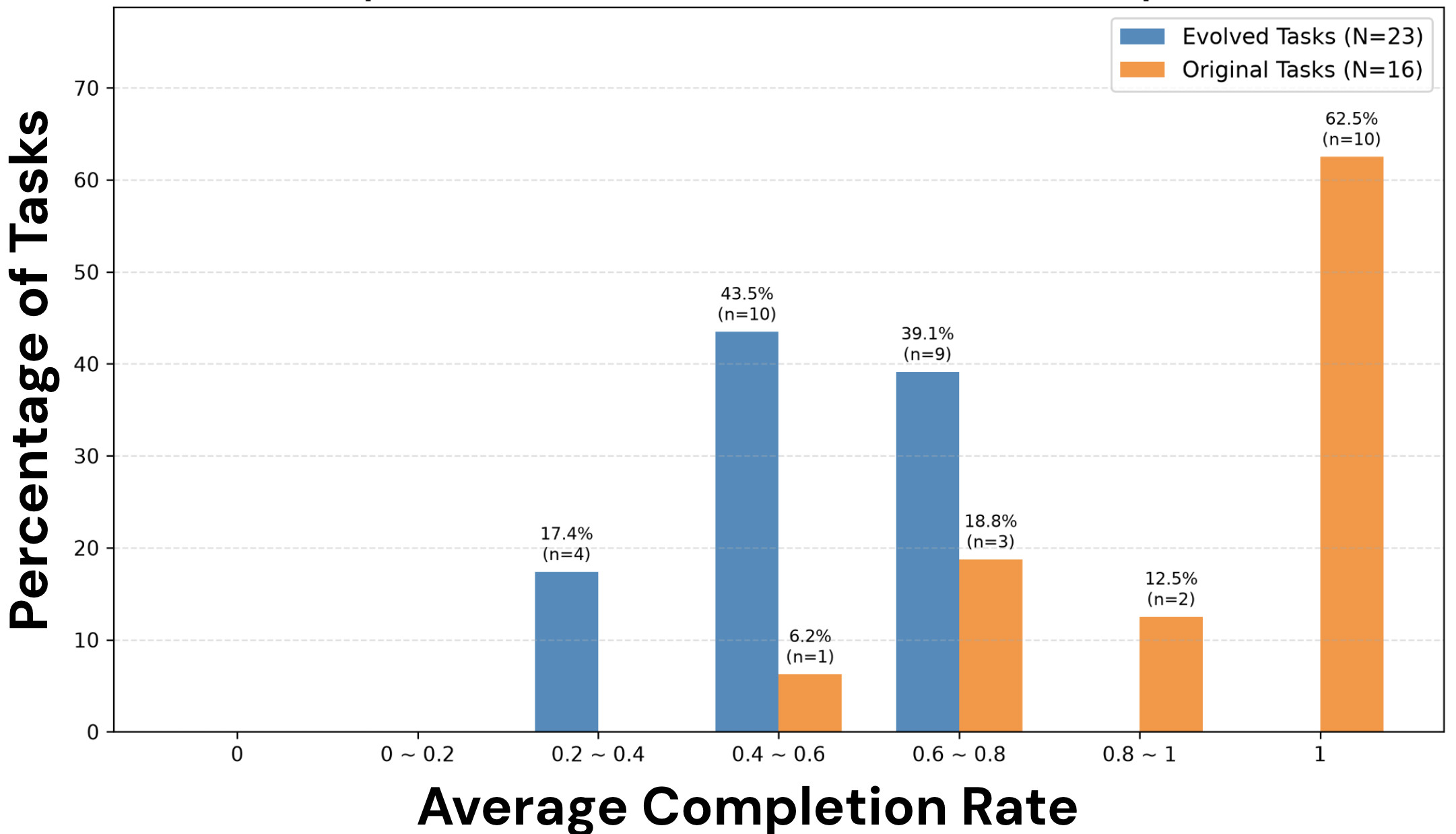}
        \caption{Average completion rate comparison of initial ($n=16$) and evolved ($n=23$) tasks in AutomationBench.}
    \end{subfigure}

    \caption{Applications of \method{} to different benchmarks with varying task difficulty levels (using the Gemma 4 31B-it solver).}
    \label{appendix:fig_task_completion_rates}
\end{figure*}

\begin{figure*}[t]
    \centering
    \includegraphics[width=\textwidth]{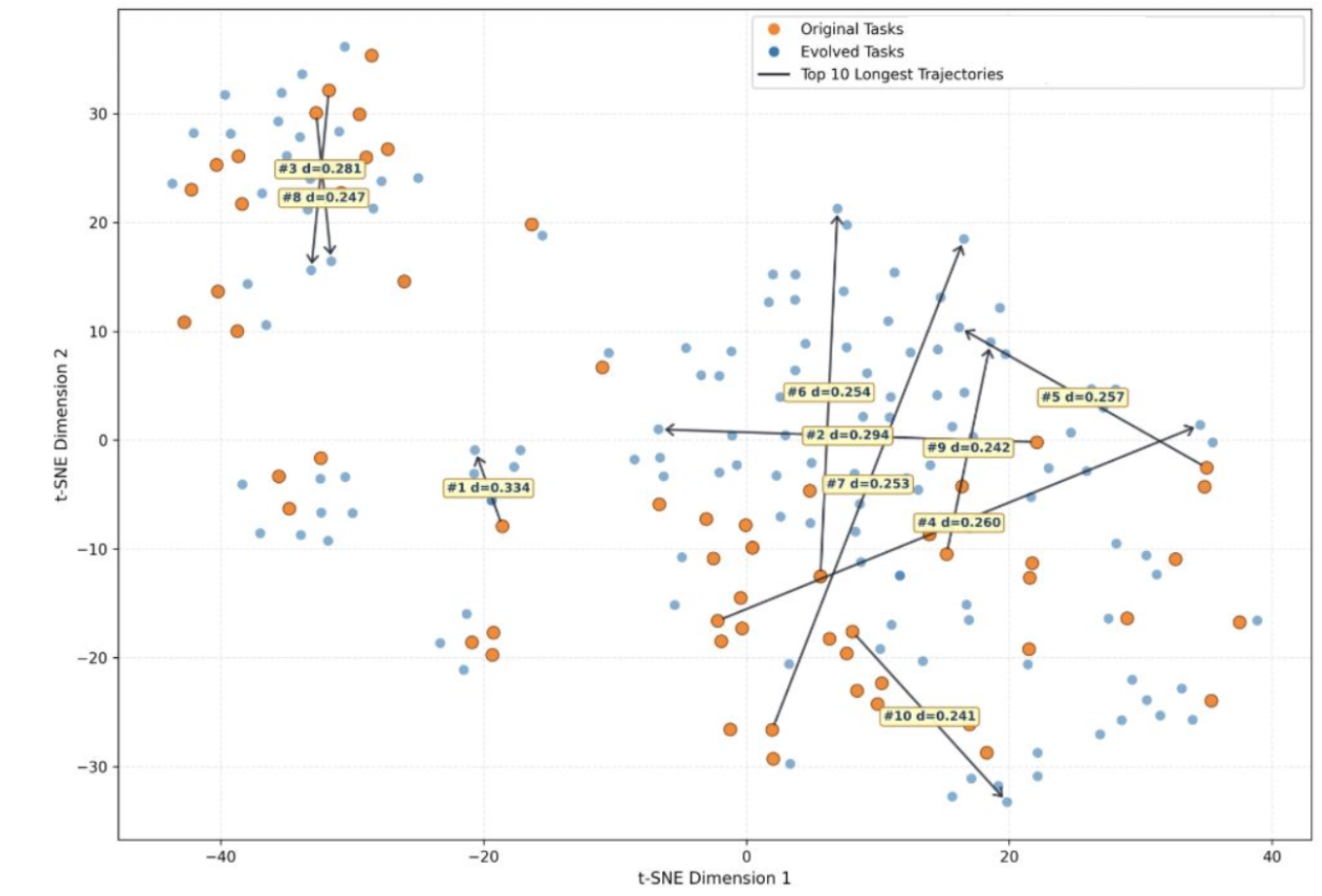}
    \caption{t-SNE visualization of the task embeddings from AppWorld using the Gemini 3.1 Flash Lite solver (original versus evolved).}
    \label{appendix:tsne_viz1}
\end{figure*}

\begin{figure*}[t]
    \centering
    \includegraphics[width=\textwidth]{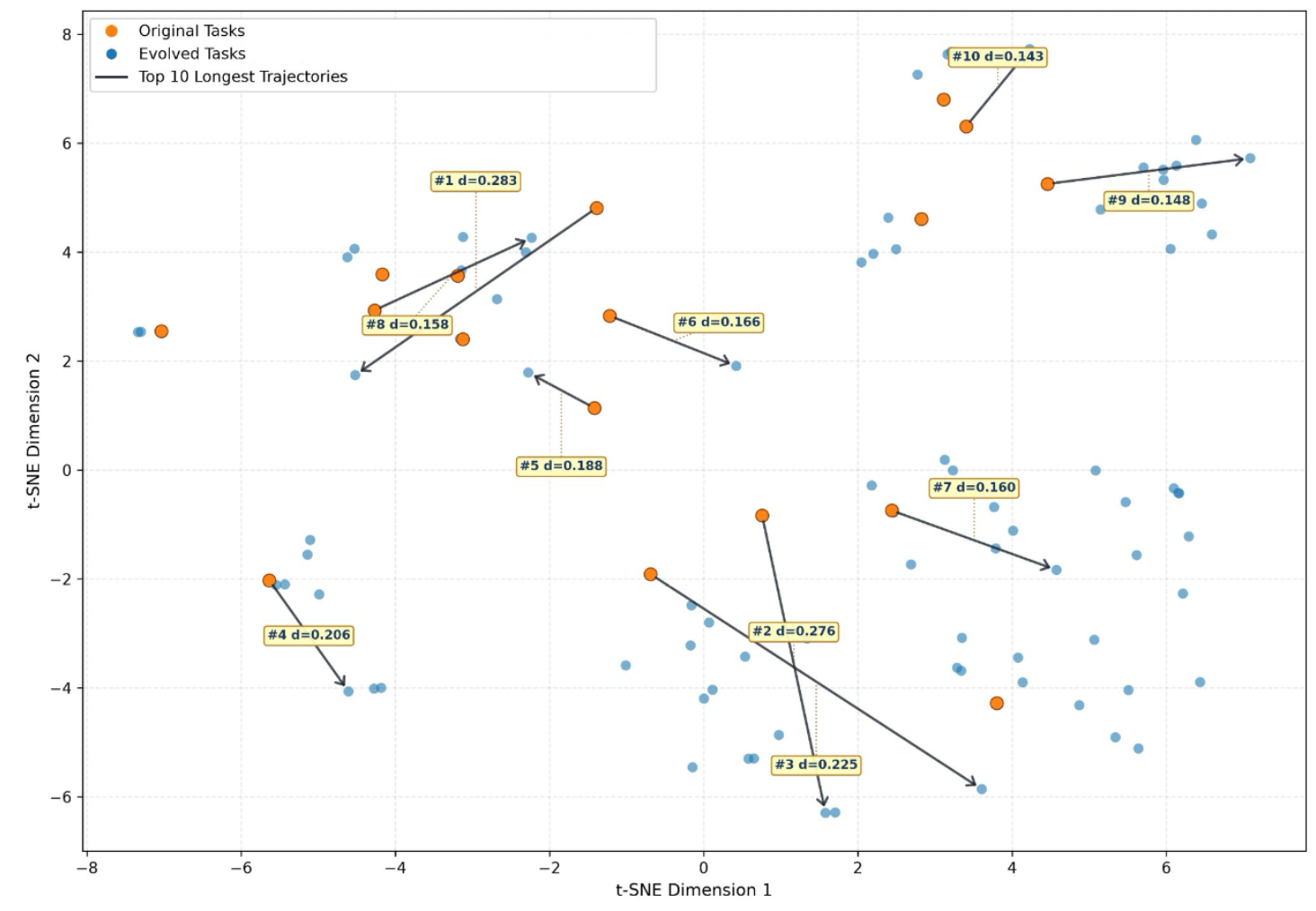}
    \caption{t-SNE visualization of the task embeddings from $\tau$-banking BM25 retrieval using the Gemma solver (original versus evolved).}
    \label{appendix:tsne_viz2}
\end{figure*}

\begin{figure*}[t]
    \centering
    \includegraphics[width=\textwidth]{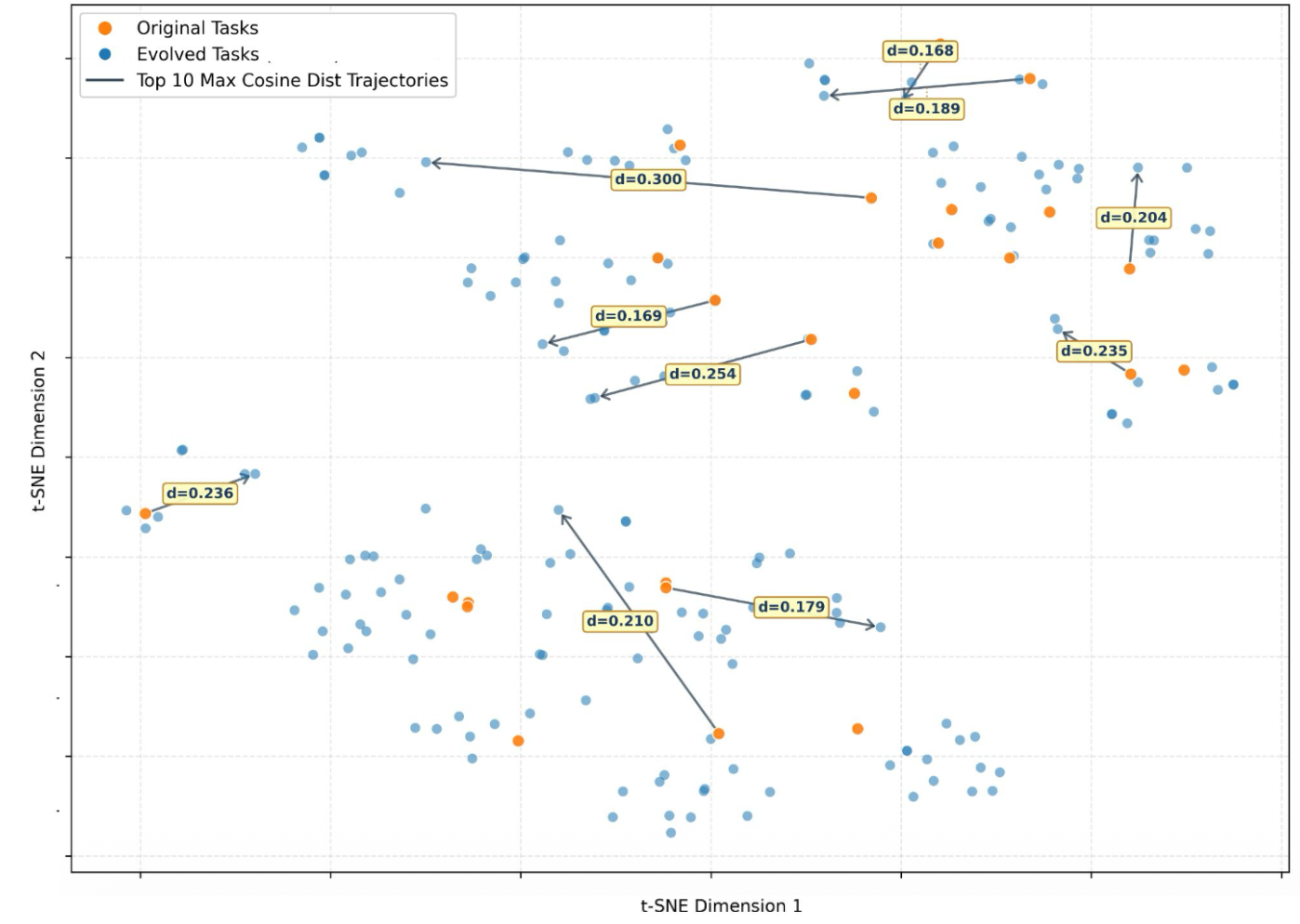}
    \caption{t-SNE visualization of the task embeddings from $\tau$-banking golden retrieval using the Gemma solver (original versus evolved).}
    \label{appendix:tsne_viz}
\end{figure*}

\begin{table*}[htbp]
    \centering
    \small
    \setlength{\tabcolsep}{4pt}
    \caption{\textbf{Taxonomy of solver failure modes.} We used Gemini 3.6 Flash as an LLM judge to identify the most likely failure mode, based on the task instruction and a summary of the solver's execution trace. This taxonomy was developed from human review of the original AppWorld benchmark, tasks generated using \method, and from error analyses provided in prior benchmarks/papers: AppWorld~\citep{trivedi2024appworldcontrollableworldapps}, AutomationBench~\citep{shepard2026automationbench}, $\tau$-banking~\citep{shi2026tauknowledgeevaluatingconversationalagents}, SWEBench~\citep{jimenez2024swebenchlanguagemodelsresolve}, GDPval~\citep{patwardhan2025gdpvalevaluatingaimodel}, and FactoryBench~\citep{merzouki2026factorybenchevaluatingindustrialmachine}.}
    \label{appendix:failure_taxonomy}
    \begin{tabular}{|p{0.16\linewidth}|p{0.14\linewidth}|p{0.28\linewidth}|p{0.34\linewidth}|}
        \hline
        \textbf{Failure type} & \textbf{Source Dataset} & \textbf{Description} & \textbf{Examples} \\ \hline
        Missing items & AppWorld & The agent failed to include or process some requested items. & If the instruction includes processing $N$ transactions, $N$ messages to friends, or $N$ downloads, the agent only processed a subset of $N$ successfully. \\  \hline
        Commonsense error & AppWorld & Non-tool-call related errors & Calculation or sorting errors (e.g., asked to identify the largest item, but the agent chose the smaller item). \\ \hline 
        Constraint violation & AppWorld, Automation Bench, $\tau$-bench, SWEBench & The agent did not respect specified constraints or priorities. & If the instruction says to only download songs by artist xyz, but songs not by this artist are downloaded; the instruction says to filter by some condition but that is not satisfied; or if the instruction has an "if-then" conditional, but the condition is not checked before executing then. \\ \hline 
        Incorrect data processing & AppWorld, GDPval & The agent processed the data incorrectly. &  Incorrectly reading the last 4 digits of a credit card number; incorrect response data parsing or extraction. \\ \hline 
        Invalid APIs & AppWorld & The agent hallucinated input, made incorrect API names, or used an incorrect API & Agent uses an incorrect tool call or the returned tool call cannot be parsed into an existing tool or API name \\ \hline 
        Incompletion & AppWorld, GDPval & The agent failed to include or process some requested items, or followed some parts of the instruction while missing other details. &  Agent returned the correct artist names but didn't download their songs, when the instruction requested both answering the artist names and downloading their songs; or adding new items to the bucket list but not marking them as incomplete; or navigating to ``photographs" folder instead of ``pictures" after making filesystem changes \\  \hline
        Incorrect search assumptions & Automation Bench, $\tau$-bench, SWEBench & If the agent's generic search fails, then they need to retrieve information differently, but the agent relies on its assumptions of where to look for things & If the data is present in notes and google spreadsheet, but the agent assumes it exists in google spreadsheet and doesn't look thoroughly across available tools \\ \hline 
        Hallucination & AppWorld & The solver made up information without retrieving this from tool calls & Making up roommate names; making up numbers or accounts \\ \hline 
        Formatting errors & AppWorld, GDPval & The agent's output formatting was incorrect. & Completed tasks correctly but made formatting mistakes in the answer, file, or message text string. \\  \hline
        Incorrect task ordering & $\tau$-bench & If either the constraint document or the logical ordering of the subtasks need to be done in a particular way, but the agent messes up the subtask order and therefore arrives at a different answer or conclusion & Subtracting then dividing is different from dividing then subtracting, leading to different outcomes \\ \hline 
    \end{tabular}
\end{table*}


\end{document}